\documentclass[a4paper,fleqn]{cas-sc}

\usepackage[numbers]{natbib}

\usepackage[para,online,flushleft]{threeparttable}

\usepackage{balance}
\usepackage{enumitem}
\usepackage[switch]{lineno}
\usepackage{caption}
\usepackage{float}

\makeatletter
\usepackage{comment}
\let\wfs@comment@comment\comment
\let\comment\@undefined
\usepackage{changes}
\let\wfs@changes@comment\comment
\let\comment\@undefined
\newcommand\comment{%
    \ifthenelse{\equal{\@currenvir}{comment}}
    {\wfs@comment@comment}
    {\wfs@changes@comment}%
}
\@namedef{Changes@AuthorColor}{red}
\definechangesauthor[name=AGS, color=blue]{AGS}
\newcommand\notsotiny{\@setfontsize\notsotiny{6.31415}{7.1828}}

\def\tsc#1{\csdef{#1}{\textsc{\lowercase{#1}}\xspace}}
\tsc{WGM}
\tsc{QE}

\begin{document}
\let\WriteBookmarks\relax
\def\floatpagepagefraction{1}
\def\textpagefraction{.001}

\shorttitle{Exploring the evolution of research topics during the COVID-19 pandemic}

\shortauthors{Invernici et al.}  

\title [mode = title]{Exploring the evolution of research topics during the COVID-19 pandemic}



%

\author[1]{Francesco Invernici}[orcid=0009-0002-5423-6978]
\ead{francesco.invernici@polimi.it}

\author[1]{Anna Bernasconi}[orcid=0000-0001-8016-5750]
\cormark[1]
\ead{anna.bernasconi@polimi.it}

\author[1]{Stefano Ceri}[orcid=0000-0003-0671-2415]
\ead{stefano.ceri@polimi.it}

\address[1]{Department of Electronics, Information, and Bioengineering, Politecnico di Milano, Milan, Italy}

\cortext[cor1]{Corresponding author}
\begin{abstract}
The COVID-19 pandemic has changed the research agendas of most scientific communities, resulting in an overwhelming production of research articles in a variety of domains, including medicine, virology, epidemiology, economy, psychology, and so on.
Several open-access corpora and literature hubs were established; among them, the COVID-19 Open Research Dataset (CORD-19) has systematically gathered scientific contributions for 2.5 years, by collecting and indexing over one million articles. 
Here, we present the CORD-19 Topic Visualizer (CORToViz), a method and associated visualization tool for inspecting the CORD-19 textual corpus of scientific abstracts.
Our method is based upon a careful selection of up-to-date technologies (including large language models), resulting in an architecture for clustering articles along orthogonal dimensions and extraction techniques for temporal topic mining. 
Topic inspection is supported by an interactive dashboard, providing fast, one-click visualization of topic contents as word clouds and topic trends as time series, equipped with easy-to-drive statistical testing for analyzing the significance of topic emergence along arbitrarily selected time windows.
The processes of data preparation and results visualization are completely general and virtually applicable to any corpus of textual documents--thus suited for effective adaptation to other contexts.
\end{abstract}



\maketitle

\section{Introduction}

COVID-19 was the first pandemic event in the Internet age. In addition to well-known social and economic implications, the worldwide community had to deal with an information overload \cite{valika2020SecondPandemicPerspective}, due to huge knowledge production about the SARS-CoV-2 virus and the associated COVID-19 disease. 
To support research, several open-access datasets, corpora, and literature hubs have been collected; among them, we mention the datasets from 
the Novel Coronavirus Information Center by Elsevier \cite{elsevierNovelCoronavirusInformation},
the National Institute of Health's iSearch COVID-19 Portfolio \cite{NIHOPAISearch},
the Human Coronaviruses Data Initiative by \texttt{lens.org} \cite{HumanCoronavirusesData},
LitCovid \cite{chen2021LitCovidOpenDatabase}, 
COVIDScholar \cite{trewartha2020covidscholar}, 
the World Health Organization COVID-19 Research Database \cite{GlobalResearchCoronavirus}, and 
the COVID-19 Open Research Dataset (CORD-19) \cite{wang2020cord}. 

CORD-19, curated by the Allen Institute for AI in collaboration with the White House Office of Science and Technology, Microsoft Research, and Kaggle, had the overall largest impact. 
Thanks to weekly updates throughout the pandemic, targeted to cover new preprints and publications, it provided a multidisciplinary, accurate, and timely view of the pandemic evolution.
The ready-to-use dataset includes curated metadata, abstracts, full-text papers, as well as vectorial representations generated by the SPECTER transformer-based language model \cite{cohan2020SPECTERDocumentlevelRepresentation}. 
At its final release, in June 2022, CORD-19 indexed more than 1 million papers (out of which 370 thousand with full text), extracted from more than 50 thousand journals and authored by more than 2 million researchers. 

CORD-19 has enabled many text mining approaches (see Wang and Lo \cite{wang2021TextMiningApproaches}), leading to remarkable results \cite{wang2021COVID19LiteratureKnowledge}, building for instance
knowledge graphs for research acceleration \cite{logette2021MachineGeneratedViewRole, wiseCOVID19KnowledgeGraph} and drug repurposing \cite{wang2021COVID19LiteratureKnowledge}, 
resource annotation services \cite{huang2020CODA19UsingNonExpert,sernagarcia2022CoVEffectInteractiveSystem},
claim verification systems \cite{wadden2020FactFictionVerifying},
and purpose-specific language models \cite{korn2021COVIDKOPIntegratingEmerging}.
Since 5 May 2023, the pandemic is no longer considered a public health emergency by the Worlds Health Organization \cite{2023WHOChiefDeclares}; 
then, we may finally consider it as a concluded phenomenon and therefore analyze its history as a whole.
In this direction, this paper aims to show how the big literature corpus CORD-19 can be successfully exploited to gather a comprehensive overview of the pandemic, tracing the trends that have characterized its scientific literature narrative.

To this end, we follow an unsupervised statistical approach based on natural language processing, specifically focused on topic modeling \cite{krause2006data}.
In the post-Large Language Models era, instead of resorting to classic topic modeling techniques like Latent Dirichlet Analysis (LDA) or Non-negative Matrix Factorization, we have chosen to exploit pre-trained language models (PLMs), providing representations that effectively embed both syntactic and semantic meaning \cite{shao2018ClinicalTextClassification}. 
PLMs can be used \textit{as is} (i.e., without any retraining) for several tasks such as summarization \cite{radfordLanguageModelsAre}, information retrieval \cite{thakur2021BEIRHeterogenousBenchmark}, and clustering \cite{reimers2019SentenceBERTSentenceEmbeddings}.

Hereby, topic modeling is interpreted as a clustering task \cite{jayabharathy2011DocumentClusteringTopic} over the latent space generated by the PLMs, as opposed to other approaches that build and train end-to-end models for topic modeling, both based on classical methods \cite{moody2016MixingDirichletTopic} and on language models \cite{meng2022TopicDiscoveryLatent}.
Along with the suggestions from the survey on topic modeling by Egger and Yu \cite{egger2022TopicModelingComparison}, we selected BERTopic \cite{grootendorst2022bertopic} to implement our analyses based on topic modeling from document clustering. 
BERTopic has already been proven a valid topic modeling framework for social sciences \cite{falkenberg2022GrowingPolarizationClimate, ebeling2022AnalysisInfluencePolitical,scepanovic2023QuantifyingImpactPositive}, since it is very flexible, can be scaled for big data corpora, and can be embedded in an end-to-end data pipeline.

In addition, we consider the textual representations extracted from TF-IDF-based models to be particularly useful and powerful when compared to both other classical methods (LDA) \cite{chen2019ExperimentalExplorationsShort} and transformer-based methods, such as Top2Vec \cite{angelov2020Top2VecDistributedRepresentations},
since it also has the advantage of becoming a knowledge retrieval proxy to discover topics by their textual representations.

Other works have previously focused on topic analysis for COVID-19-related matters;
some analyzed the early stages of the pandemic \cite{zhang2021TopicEvolutionDisruption,tran2020StudiesNovelCoronavirus}, 
others analyzed the broader field of coronaviruses \cite{pourhatami2021MappingIntellectualStructure}, 
focused on topic distribution by country \cite{berchialla2021EffectCOVID19Scientific} or on the delineation and impact in scientometric terms of the early CORD-19 \cite{colavizza2021ScientometricOverviewCORD19}.
The approach conducted in this study,
hereon called CORToViz (CORD-19 Topic Vizualizer)
is broader, as it applies to the entire pandemic history without choosing a specific field of investigation \textit{a priori}.

As our input, we consider all the English-language abstracts of CORD-19 with high-quality metadata that were published after December 2019.
First, we provide a pipeline for ingesting huge data corpora, built upon state-of-the-art technologies \cite{grootendorst2022bertopic}, and extracting from them highly relevant topics, clustered along orthogonal dimensions.
Then, our system enables the discovery, from a given literary corpus, of topics of interest through a keyword-based search interface. For the discovered topics, a word cloud representation is rendered to the user, who can select, based on the insights of the content, 
a set of topics whose trends should be visualized on the timeline of the pandemic. When any topic presents evident trends for distinct time periods, a statistical test can be run to determine if it is a significant behavior or just a stochastic event. Our approach leverages the existing technology of BERTopic exclusively for topic analysis, as results are further elaborated by binning topic-clustered documents within temporal ranges, then obtaining relative bin-representation frequency, and finally producing interactive statistical testing. 

Remarkably, CORToViz enables a fast exploratory analysis of a big data corpus along the time dimension, in a way that was not well supported previously. Additionally, the proposed technology and method are completely general and agnostic to the specific domain; our full-stack process is applicable to any corpus of medium-sized textual documents, using any topic model of choice, and a time-series visualizer.
We foresee that a methodology similar to the one presented here, once deployed on the Web, can support the lightweight analytics of arbitrary domains.

\section{Methods}

\subsection*{CORD-19 anatomy}

\begin{figure}[h!]
\centering
    \includegraphics[width=\linewidth]{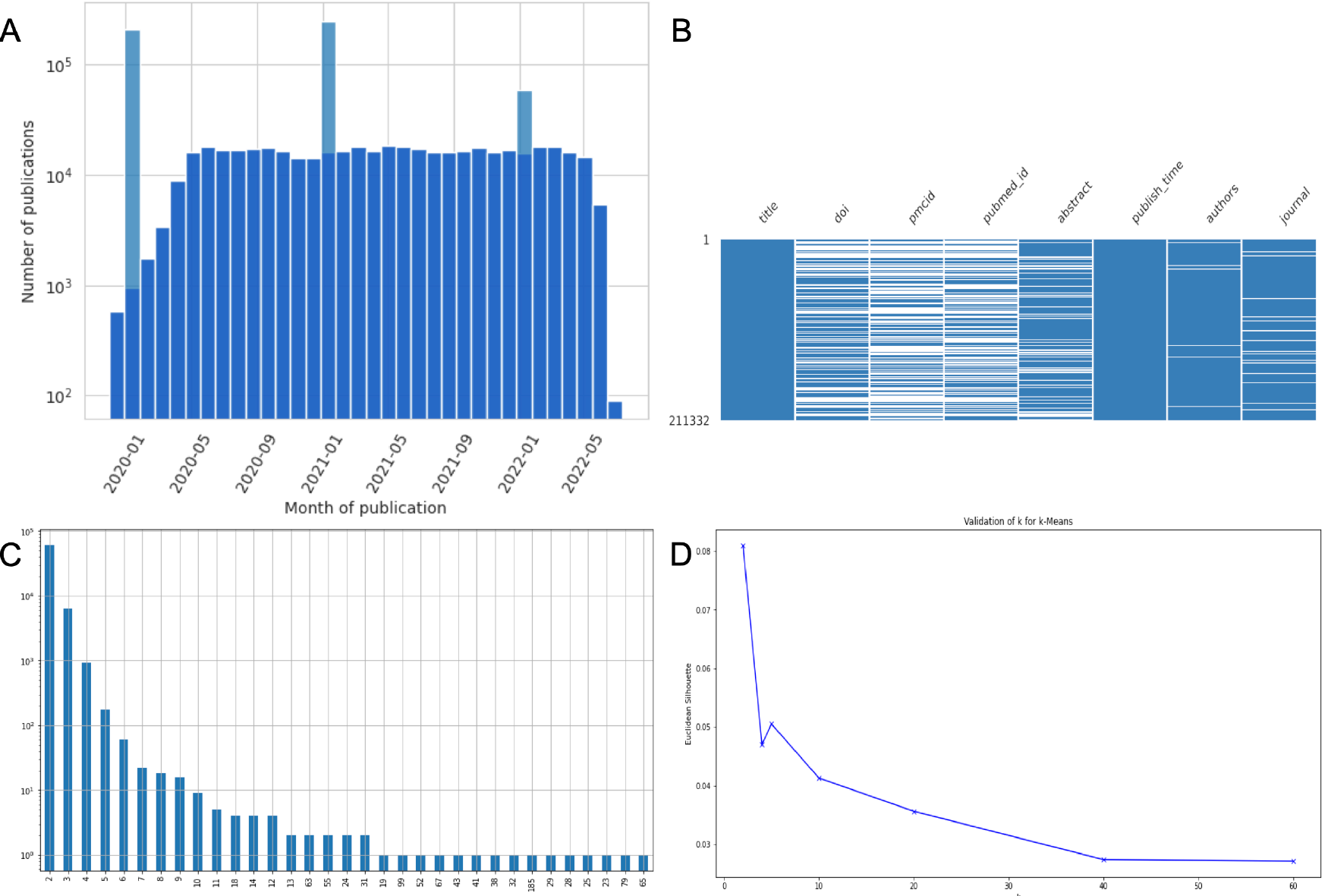}
    \caption{\textbf{Visualizations of the exploratory analyis of CORD-19 data and metadata.}
    (A) Monthly number of publications in CORD-19. The number increases in the first months of 2020, then is rather stable, until April 2022, when the trend started decreasing; CORD-19 was updated until June 2022. In light color, the spikes of publications with just the year in their metadata were converted to the first of January; these entries were removed.
    (B) Data-density display of eight metadata fields for a sample of 20\% of the dataset. We retain articles with abstract and publish\_time metadata.
    (C) Distribution of the number of duplicates. The majority of articles, on the left of the distribution, have a single duplicate (typically without the doi), representing a preprint non-peer-reviewed version uploaded on public archives before publication; only a few documents are present in the dataset with a high number of replicated entries.
    (D) Silhouette score for k-Means for different values of k, which indicates the number of clusters. A spike in the line plot means that that value is a good candidate for the number of clusters; the figure clearly indicates that five is a good candidate, then selected for the exploratory clustering analysis.}
    \label{fig:panel_eda}
\end{figure}

The content of CORD-19 has been explored with an in-depth analysis of its metadata and embeddings \cite{cohan2020SPECTERDocumentlevelRepresentation}.
CORD-19 presents 1,056,660 articles with associated metadata.  Fig.~\ref{fig:panel_eda}\textit{A} presents their distribution along their month of publication, showing a rapid increase in scientific production from the pandemic outbreak until the summer of 2020, and then a stable production. Metadata is typically incomplete, due to missing entries in several fields; this can be observed in  Fig.~\ref{fig:panel_eda}\textit{B}, where we show eight metadata fields for a representative 20\% of the dataset.
Several papers come with duplicates -- as analyzed in  Fig.~\ref{fig:panel_eda}\textit{C}; in such cases, we retained only the most representative paper of duplicate clusters (see details in the \textit{Materials and Methods}). After data filtering, we retained 357,170 distinct abstracts written in English that presented adequate metadata for our research purposes.
We only selected abstracts equipped with publish\_time information (see Fig.~\ref{fig:panel_eda}\textit{B}).

\subsection*{Preliminary and fine-grain clustering}

\begin{figure}[h!]
    \centering
    \includegraphics[width=0.96\linewidth]{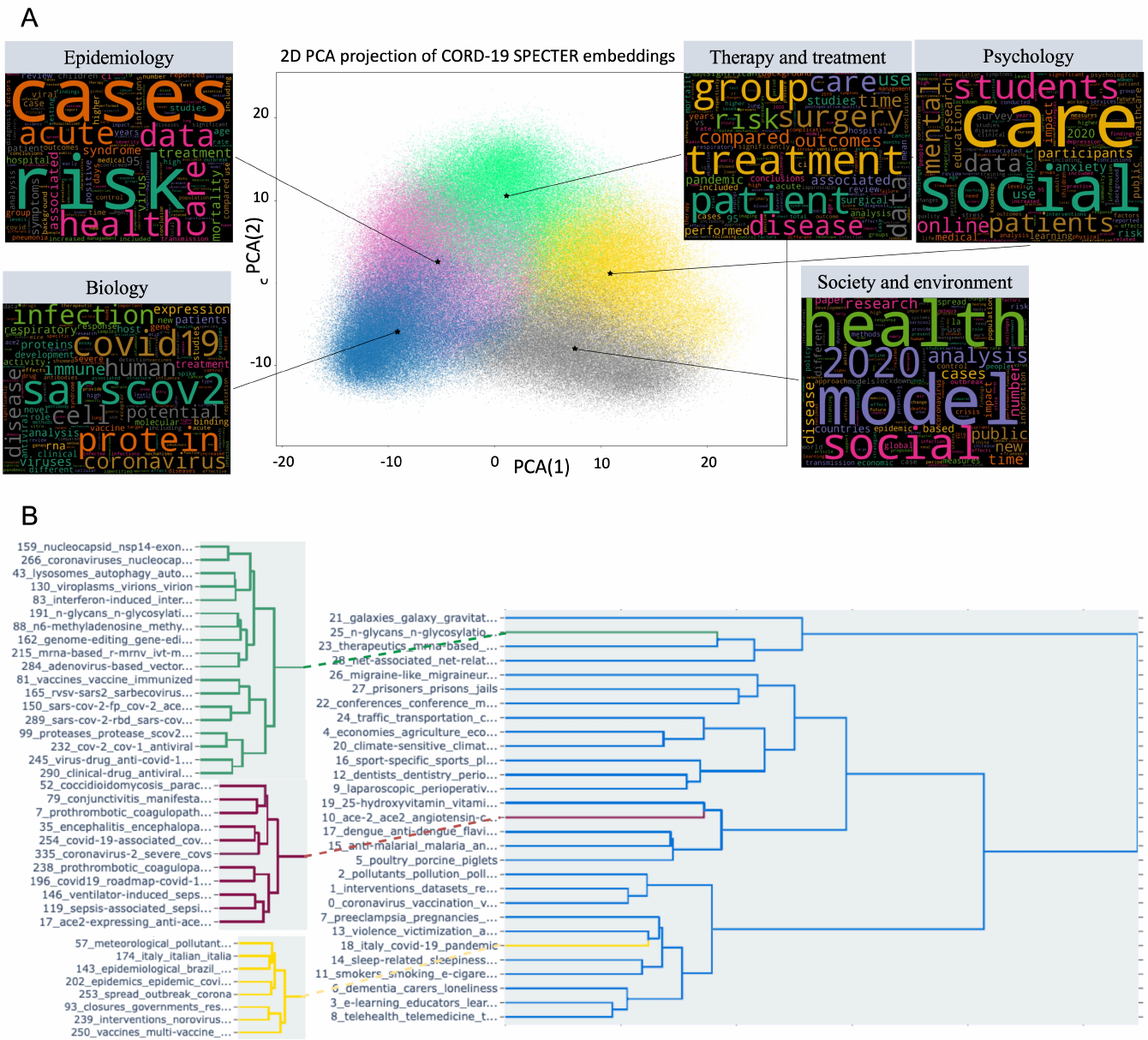}
    \caption{{Topic clustering produced by the preliminary and fine-grain clustering methods.}
    We show and compare the clusters introduced in the Results section. 
    (A) Scatter plot of the exploratory clustering analysis.
    The analysis has been performed with k-Means, a classic clustering algorithm. We found five macro-topics and we assessed their content with word clouds. As shown in the figure, the five clusters identify distinct classes of topics, well described by word clouds, which nicely partition the set of articles of CORD-19.  
    (B) Dendrograms of the hierarchical density-based clustering.
    We then explored topics using a technology-rich pipeline, resulting in a fine-grain topic clustering. The high-level cluster hierarchy, with only 29 clusters, resembles the five macro-topics structure of the preliminary clustering. The full hierarchy includes 354 fine-grained clusters, each related to a specific high-level cluster. We show the hierarchy of the \emph{n-glycans}-related topics, of the \emph{ACE2}-related topics, and of an epidemiology-related topic.} 
   \label{fig:clusters}
\end{figure}

Then, we performed a preliminary feasibility assessment of the topic modeling analysis. We aimed to verify if the latent topics’ structure of CORD-19 can be modeled as an unsupervised clustering task.
First, we selected the optimal value for executing k-Means clustering (see Fig.~\ref{fig:panel_eda}\textit{D}). 
As a result of this exploration, we obtained a comprehensive view of the corpus, split into five clusters, reported as a scatter plot in Fig.~\ref{fig:clusters}\textit{A}. Thanks to the word clouds generated from the most frequent words for each cluster, we were able to identify five macro-topics:
\begin{enumerate}[noitemsep]
\item {\it Biology of coronavirus}, associated with words `protein', `cell', and `viral', with 68,278 articles.
\item {\it Therapy and treatment}, associated with words `group', `treatment', and `patient', with 54,391 articles.
\item {\it Epidemiology}, associated with words `cases', `data', and `risk', with 99,554 articles.
\item {\it Psychology}, associated with words `social', `mental', `care', and 'students', with 72,980 articles.
\item {\it Society}, associated with `model'/`social' accompanied by `analysis', `research', `public', and `lockdown', denoting a broad cluster on the impact of the pandemic on society, with 61,967 articles.
\end{enumerate}

For our purposes, however, we needed a much finer grain of the topics' structure, in order to best identify and track specific phenomena that characterized the pandemic. 
Thus, we ran a technology-rich, state-of-the-art pipeline, detailed  next,
obtaining a considerably richer topics' structure. In particular, given the hierarchical nature of the adopted algorithm, we obtained a hierarchy of 354 clusters, each of which defines one topic; 
finely-grained topics are aggregated as a list of 29 high-level topics, shown in Fig.~\ref{fig:clusters}\textit{B}; these topics, while not perfectly overlapping, can be related to the macro-topics of the exploratory analysis. For instance: 
(1) the high-level cluster on n-glycans, at a finer-grain level also includes clusters on the nucleocapsid, lysosomes, and methyladenosine relates to the ‘biology’ macro-topic;
(2) the high-level cluster covering ACE2 and angiotensin,  is expanded by clusters on prothrombic coagulopathies, COVID-19-associated conditions, severe Coronavirus Disease, sepsis, and ventilators, relates to the ‘therapy and treatement’ aspects;
(3) fine-grain clusters related to Italy’s COVID-19 pandemic, outbreak, government restrictions, and immunization relate to the ‘epidemiological’ macro-topic; 
(4) clusters on e-learning, caretakers, loneliness, and sleep-related issues relate to the ‘psychology’ macro-topic; and 
(5) clusters on traffic, transportation, pollution, and economy are related to the ‘society and environment’ macro-topic.

\subsection*{Architecture for topic extraction and dynamic modeling}

\begin{figure}[h!]
    \centering
    \includegraphics[width=\linewidth]{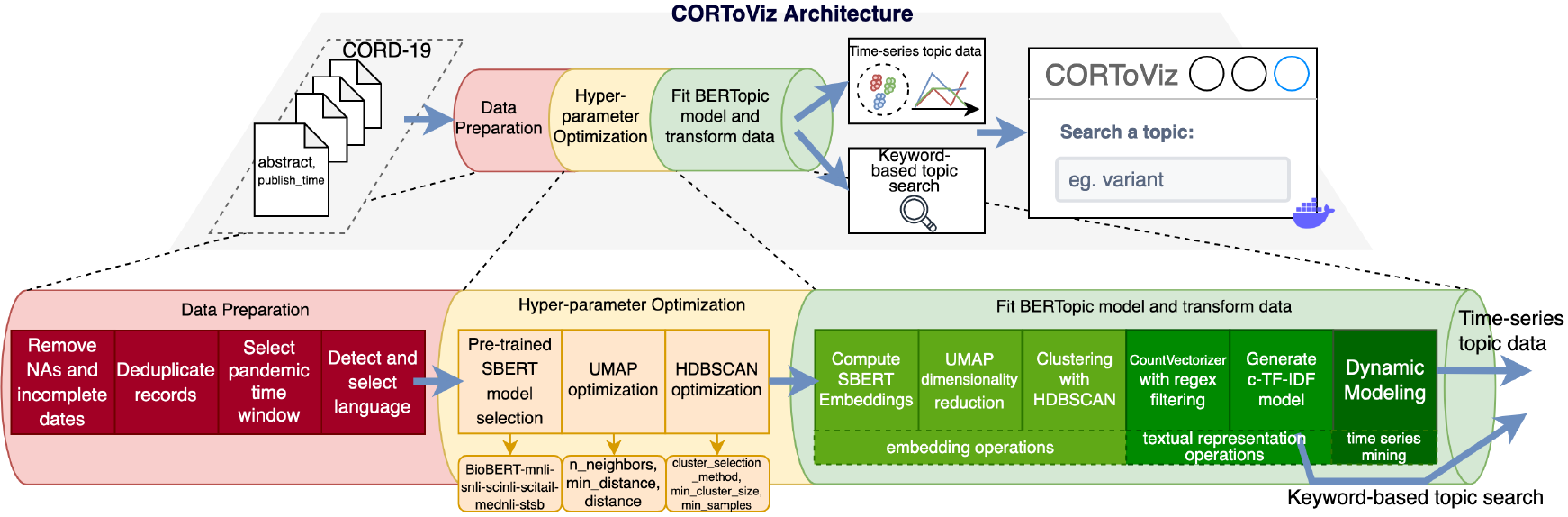}
    \caption{{General architecture of CORToViz.} 
    The data pipeline consists of three stages: data preparation (red), hyperparameter optimization (yellow), and topic extraction using the BERTopic model (green); the pipeline produces as output the ingredients for the dashboard application, a user-friendly interface for topic selection and display. 
    In the data pipeline, the data preparation step selects the abstracts with the appropriate metadata from CORD-19; the hyper-parameter optimization finds the values that maximize the performance of the models operating on embeddings; finally, the data transformation generates the artifacts used by the CORToViz dashboard application. The dashboard supports keyword-based topic search and then visualizes the time series information for each topic; each topic is associated with a word cloud, providing insight into the topic's content.}
    \label{fig:architecture}
\end{figure}

To process the original dataset, extract topics, and prepare their time-series data, we adopted a full-fledged, technology-rich complete pipeline, illustrated in Fig.~\ref{fig:architecture}.
The pipeline assembles up-to-date technologies and is fully portable, after adaptation, to any organized repository such as CORD-19.

The top subfigure provides a bird-eye-view of the pipeline.
The `Data Preparation' phase is in charge of removing records with null values or incomplete dates, retaining only records referring to articles written in English and published between December 2019 and June 2022. Record deduplication is also performed since -- especially during the pandemic -- several scientific contributions were exposed through different portals and registered with multiple digital object identifiers.
The `Hyper-parameter Optimization' phase 
selects the specific large language model instance and optimizes parameters for dimensionality reduction and topic learning. 
Finally, the `Fit BERTopic model and transform data' phase includes 
a) embedding operations (executing a dimensionality reduction and density-based clustering -- with optimized parameters); b) textual representation operations; and c) time series mining.
The processes in (a) and (b) enable the keyword-based topic search, whereas (c) enables plotting time series of topic data in the visualization tool.
The next paragraphs describe the pipeline steps more in detail.

\subsubsection*{Metadata selection}
The pipeline ingests the CORD-19's metadata table, and applies several preprocessing operations, as shown in the ``Data Preparation" section in Fig.~\ref{fig:architecture}; each row of the metadata table (1,056,660 records) corresponds to a distinct document of the CORD-19 collection (e.g., an abstract, an article, and so on). 
The pipeline filters rows with {\it missing mandatory information}, such as the title, abstract DOI, and publishing time; in particular, rows with incomplete time (e.g., with just the publishing year) were also filtered. Next, records were {\it deduplicated}, thanks to the {\tt cord\_uid}, linked with groups of CORD-19 entries describing the same paper. Most CORD-19 deduplication refers to differences in external identifiers, while duplicates agree on textual metadata, the focus of this work. Finally, only documents written in English within the time window between December 2019 and June 2022 were selected, yielding a final collection of 327K entries.

\subsubsection*{Terminology}
In this research, several methods and technologies are derived from different domains (i.e., language models, clustering, vectorization, topic modeling, and search), each using specific terms. Below, we clarify how such terms relate to each other.
\begin{itemize}
    \item {\bf Token, word, term.}
    Scientific abstracts are composed of \textit{words}, whose meaning is dictated by common sense;
    language models transform words into \textit{tokens}; vectorization methods associate frequencies to each \textit{token}, and these are used for building word clouds.
    Finally, search systems
    denote as \textit{terms} the words used in keyword-based methods (e.g. TF-IDF) and search.  
    \item {\bf Cluster, class, topic}.
    Once abstracts are represented as points in the space of the embeddings, they form {\it clusters}, or {\it classes} of similar elements; they are then called \textit{topics} after being associated with a textual representation understandable by humans (e.g., a word cloud).

\end{itemize}

\subsubsection*{Unsupervised topic modeling}
Unsupervised Topic Modeling is used to discover and analyze latent topics within a document, without pre-existing labels or supervision. The methods work at best under the assumption that each document represents a single topic, or at least that one topic is preponderant, so as to exclude featuring multiple topics at the same time. We applied unsupervised topic modeling to CORD-19 abstracts; we based this work on BERTopic \cite{grootendorst2022bertopic}, a topic modeling tool that leverages transformers and clustering models for latent topic identification and a class-based term frequency–inverse document frequency model for textual representation learning.
We produced as output the topics' identification, their representations as word clouds, and the temporal distribution of topics over time. In the lack of ground truth, we employed quantitative methods, wherever possible, to assess intermediate models.

\subsubsection*{Preliminary feasibility assessment}
At first, we tested the feasibility of learning the latent topics' structure of CORD-19 as an unsupervised clustering task, by applying the K-means clustering algorithm to embeddings of the dataset texts. Embeddings were generated by SPECTER \cite{cohan2020SPECTERDocumentlevelRepresentation}, a generic multi-task document-level transformer model. 
To select the optimal number \textit{k} of clusters, we maximized (in the 2-60 range) the silhouette score \cite{shahapure2020ClusterQualityAnalysis} over a random sample that covers 25\% of the dataset, as shown in Fig. \ref{fig:panel_eda}\textit{D}. In this way, we identified five macro-clusters. To assess the content of these macro-topics, we computed the 100 most frequent words present in the abstracts (after stopword removal), by using \texttt{gensim} \cite{rehurek10softwareframworktopic}, a Python package for topic modeling. Then, we generated a word cloud for each cluster; words are displayed in different sizes based on word frequency. In this way, we determine that clusters cover different areas of research, as reported in Fig. \ref{fig:clusters}, generated by projecting the embeddings on the first two principal components \cite{james2021unsupervised}.

\subsubsection*{Learning the best latent representation}

\sloppy{In order to learn the latent topic structure of the dataset, we mapped each abstract from CORD-19 to a point in an embedding representation, consisting of a dense, 768-dimensional, vector space. Vectorial representations were generated by a transformer model, selected among the sentence transformer models \cite{reimers2019SentenceBERTSentenceEmbeddings} compatible with BERTopic. 
Specifically, we used from the \texttt{huggingface} repository the model \texttt{pritamdeka/BioBERT-mnli-snli-scinli-scitail-mednli-stsb} \cite{deka2022EvidenceExtractionValidate}, which provides robust sentence embeddings for clustering and information retrieval tasks for scientific and medical literature.} 

Thus, we modeled the latent topic learning problem as a clustering task. We adopted HDBSCAN \cite{mcinnes2017HdbscanHierarchicalDensity} because it is a density-based clustering algorithm. This characteristic helps to learn clusters that are not perfectly shaped as hyperspheres. 
HDBSCAN is also fairly tunable, to avoid a cluster structure with degenerate characteristics, such as a single massive cluster surrounded by multiple single-item outlier clusters.
In order to improve the quality of clustering we compressed the embeddings with UMAP \cite{mcinnes2020UMAPUniformManifold}, which is a stochastic dimensionality reduction algorithm, before feeding them to HDBSCAN. By tuning UMAP, we found a sweet spot in the trade-off between keeping local structures and prioritizing the representation of the global structure.

To select the hyperparameters of the UMAP and HDBSCAN models, we performed an optimization step, by grid searching the values on the two jointed models targeting the highest DBCV \cite{moulavi2014DensityBasedClusteringValidation} value, which is a score for the goodness of a density-based clustering model spanning [-1, 1]. At every iteration of the grid search, we randomly sampled 25\% of the abstracts from the data preparation. In the best run, we obtained a DBCV score of 0.36, with the following values as parameters of UMAP: 
\texttt{n\_neighbors = 50}, 
\texttt{n\_components = 50}, 
\texttt{min\_dist = 0.0}, and
\texttt{metric = `cosine'}, 
and with the following values as parameters of HDBSCAN: \texttt{min\_cluster\_size = 100}, 
\texttt{min\_samples = 10},
\texttt{metrix = `euclidean'}, and
\texttt{cluster\_selection\_method = `leaf'}.

\subsubsection*{Extraction of textual and visual representations for topics}
After we found clusters in the latent space of embeddings, we searched a synthetic representation of each cluster, for understanding the content of the abstracts and finally defining topics. Again, we adopted the stack proposed by Grootendorst \cite{grootendorst2022bertopic}, which is available in BERTopic. As it can be appreciated in Fig. \ref{fig:architecture}, the task consists of two steps: 1) abstract vectorization and 
2) fitting of per-class TF-IDF \cite{ceri2013InformationRetrievalModels} models. 
We used the \texttt{scikit-learn} \cite{pedregosaScikitlearnMachineLearning} CountVectorizer, which
converts a collection of text documents to a matrix of token counts; we set
\texttt{stop\_words} to ``english" and \texttt{token\_pattern} as a regular expression to keep together hyphenated words, such as COVID-19 and SARS-CoV-2, which are common in biomedical writings. Similarly, we fitted the c-TF-IDF model with the \texttt{reduce\_frequent\_words} parameter set, which considers the square root of the normalized frequency of the terms (i.e., words). With this model, we obtained -for each class- the most relevant terms (i.e., topics) and their frequency. In this way, we computed a textual, human-understandable representation for each cluster, and then retrieved the most important topics using the TF-IDF representations. Finally, we adopted the \texttt{word cloud} \cite{mueller2023Wordcloud} package to generate word clouds with the most frequent terms of each topic, thereby providing a visual representation to inspect the content of the topic.

\subsubsection*{Dynamic topic modeling}
To understand the trends of the research topics during the pandemic, we used the dynamic modeling tool of BERTopic.
In this tool, the classical definition of topic modeling is extended by including the temporal dimension; to do so, Grootendorst \cite{grootendorst2022bertopic} employs the concept of absolute counts for each topic in equally-sized temporal bins, whose size in days has to be defined before the extraction of the time series with such data.
In this work, we build on this idea, but we decided to use relative frequencies, by normalizing each absolute count of topic observations with the amount of abstracts published in that period. In this way, we can interpret these values as pointwise measures of the intensities of the topic, as other previous works on dynamic topic modeling \cite{krause2006data}.
In practice, we extract the absolute frequency of each topic within time bins of equal size, specifically 1, 2, 3, and 4 weeks; we paired the abstracts with their publication date (\texttt{publish\_time}). We then pivoted the resulting data frames in order to obtain actual time-series data for each topic. 
We also normalized each row of the pivoted dataframe, representing a time bin, to obtain the relative frequency of each topic for that time frame. Taking advantage of these time-series, we generated line plots and stacked histograms for the counts of abstracts per bin. 
To put these visualizations in chronological context, we added as background the plot of the global number of active COVID-19 cases, retrieved from Our World In Data \cite{mathieu2020CoronavirusPandemicCOVID19}, and a timeline of significant events that marked the evolution of the pandemic, such as lockdowns, vaccination campaigns, and variants' outbreak, collected from various resources \cite{COVID19SARSCoV2Coronavirus, InformationCOVID19, 2023TimelineCOVID19Pandemic}.

\subsubsection*{Statistical evidence for topic dynamics}
To check if a topic's trend is statistically significant, we used a procedure based on the non-parametric Kruskal-Wallis test \cite{kruskal1952UseRanksOneCriterion}, used for comparing sample medians, checking if two groups are sampled from the same population. 
The test can be conveniently parametrized by choosing on which topic and time frames (T1 and T2) it should be run;
starting and ending times of time frames T1/T2 can be set.
The test produces p-value and H statistics, enabling the acceptance or rejection of the simple null hypothesis, which corresponds to ``there is no significant difference in the topic representation in periods T1 versus T2".
Specifically, we adopted the implementation of the test available in the Python library \texttt{SciPy.stats} \cite{2020SciPy-NMeth}, which implements the formulation of the H-test statistic with correction for ties (i.e., two observations among all the groups are equal).

At first, each observation is assigned a rank, starting from 1 for the lowest value. If there are ties, each observation is given the mean of the ranks for which it is tied. Then, the H statistic is computed as

\begin{equation}
H = \frac{\frac{12}{N(N+1)}(\frac{R_1^2}{n_1} + \frac{R_2^2}{n^2}) - 3(N+1)}{1-\frac{\sum T}{N^3-N}}
\end{equation}

where $n_1$ and $n_2$ are the numbers of observations in the two groups,  $R_1$ and $R_2$ are the sums of the ranks of the observations of the two groups, $N$ is the number of total of observations of the two groups combined and the summation of $T$, where $T = (t-1)t(t+1)$, is over all the groups of ties.
The H test statistic takes positive values and the critical value for the 5\% p-value, which we use as the threshold for significance, for two groups, is 3.85. Higher values imply lower p-values and, hence, the rejection of the null hypothesis \cite{kruskal1952UseRanksOneCriterion}.

\subsubsection*{An interactive, discovery-enabling dashboard}
All the components described before, such as the c-TF-IDF model, the time-series plots, the word clouds, and the statistical box, are embedded in a single-page, interactive, and responsive dashboard.
Users can interactively set the topic and time frames to be tested, therefore impacting the selection of considered abstracts;
they can visualize the p-value and H statistics, easily testing their own hypotheses and drawing conclusions.

For implementing the interface, we used  \texttt{Streamlit} \cite{2021StreamlitFasterWay}, a Python package for building single-script web applications. In this way, we enabled multiple users to explore the topics of CORD-19 at the same time. The application, with the data and the model, has been dockerized to facilitate distribution and deployment.

\subsubsection*{Building document corpora for alternative domains}
We implemented a method to build another corpus of abstracts from scientific literature exploiting the public endpoint of the APIs provided by the Springer Nature Group. Specifically, we requested from the Springer Meta API (\url{http://api.springernature.com/meta/v2/json}) all the articles listed under the subject ``Climate Change'' 
(specified in the \texttt{querystring} parameter \texttt{q} of the endpoint \cite{nature2023api}). 
Such a query allowed us to retrieve a dataset of 33723 abstracts.
We then 
cleaned the dataset through the Data Preparation stage of the CORToViz pipeline, 
obtaining the final dataset of abstracts;
the following steps of the pipeline (Fig.~\ref{fig:architecture}) were then applied to obtain a topic model and a dataset suitable for exploration using our topic search dashboard.


\section*{Results}

\subsection*{Topic visualizer}

\begin{figure}[h!]
\centering
    \includegraphics[width=0.9\linewidth]{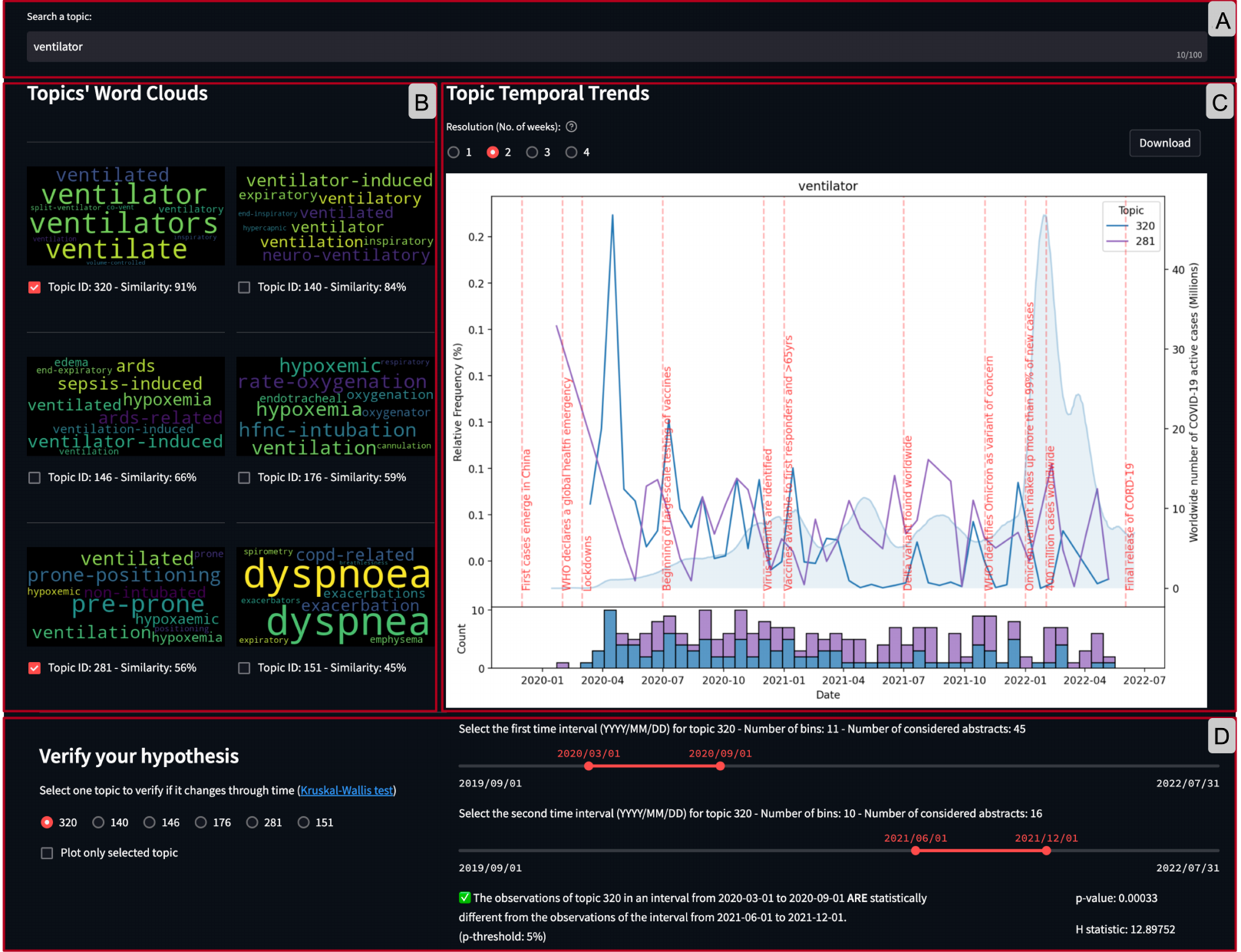}
    \caption{{User interface of the CORToViz dashboard.}
    (A) Keyword-based search bar - the example query ``\textit{ventilator}" is entered by the user.
    (B) Six top-ranked topics, explained through their word clouds. The user selects two topics ``{\it ventilator}" and ``{\it prone-positioning}". 
    (C) Line plot of the 
     intensities (i.e., the relative frequencies of appearance) of the two selected topics. The user sets (above) the bin resolution to 2 weeks (options are 1-4 weeks). Histograms (below) show the count of articles associated with the selected topics, with the given bin size of two weeks. 
    (D) Panel showing statistical testing. The user selects the "{\it ventilator}" topic and sets two time windows, a six-month window at the beginning of the pandemic and a 6-month window at the end of the second year of the pandemic. The tool, at the bottom, reports the result of the Kruskal-Wallis test for the difference between groups. Specifically, it shows the H statistic of 12.89752, which is the statistic of the aforementioned non-parametric test, and that determines the p-value (0.00033). Therefore, since the p-value is below the 5\% threshold, the null hypothesis (i.e., no difference in groups) can be rejected, and a green check indicates a statistically significant difference between observations in the two intervals.}
    \label{fig:interface} 
\end{figure}

To support the full exploration of the fine-grain topic structure of CORD-19, we developed the Topic Visualizer, available as an interactive dashboard (\url{http://gmql.eu/cortoviz/}).
The tool includes several interfaces for topics exploration, tracking their evolution in time, and associating behavior with their statistical significance. Users will experience a free keyword-based topic search, producing -- as an outcome -- results in the format displayed in Fig.~\ref{fig:interface} within a few seconds.

Fig.~\ref{fig:interface} portrays the four main areas of the dashboard.
Panel (A) shows the search feature, where users can enter an arbitrary search keyword to start their search session. To demonstrate the approach, Fig.~\ref{fig:interface}\textit{A} shows the search of topics related to `ventilator'. 

Panel (B) shows six top-ranked topics through their word clouds, ordered by similarity (spanning from 0.91 for topic ID 320 to 0.45 for topic 151). The user can filter topics for further visualization and analysis (in the example, topics `ventilator' and `prone-positioning' are selected). 

Panel (C) shows the plots of the time series of articles associated with the selected topics, respectively through line plots (above) and stacked histograms (below). The y-axis of the line plot shows the amount of scientific abstracts represented in the shown topic using their relative frequency w.r.t. to the total number of abstracts for each bin, whereas the y-axis of the histogram holds absolute counts of scientific papers for each bin. The relative frequencies of appearance for determined periods represent pointwise measures of the topic intensity \cite{krause2006data}. The x-axis is used to scan the temporal evolution, binned according to time resolution, ranging between one and four weeks, that can be manually changed. 
Plots are augmented with contextualizing statistics: a plot of global active cases of COVID-19 (expressed in million units and highlighted using a gray area underneath the plot) and vertical red lines marking peculiar events of the pandemic (first cases in China, the world's emergency declaration, first lockdowns, start of vaccine testing, start of vaccination campaign, onset of Delta and Omicron variants, worldwide peak of 400 million cases, final release of CORD-19).

Panel (D) shows the box for testing the statistical significance of changes in topic intensity.
Users specify the topic to be tested among those shown in panel (B). The user also uses sliders to define two time intervals. All articles whose publishing date falls within the two intervals are included within two groups, 
and the Kruskal-Wallis test -- a non-parametric test on the difference of the medians of intensity values -- is performed; the null hypothesis states that there are no significant differences among the groups, while the alternative hypothesis suggests that the two groups differ. 
The null hypothesis is rejected when the test's H statistic is above a certain threshold (correspondingly, the test's p-value is below a certain threshold, i.e., 5\%).
Fig.~\ref{fig:interface}\textit{D} shows that the topic with ID 320 (focused on different formulations of ventilator/ventilation) was significantly different (indeed, much more intense) in the March-September 2020 interval when compared with the June-December 2021 interval, having a p-value below 0.03\%; when the difference is not significant, a red cross appears (see \textit{Materials and Methods} for details on the H statistic of the test).

\subsection*{Search sessions}

\begin{figure}[h!]  
    \centering
    \includegraphics[width=\textwidth]{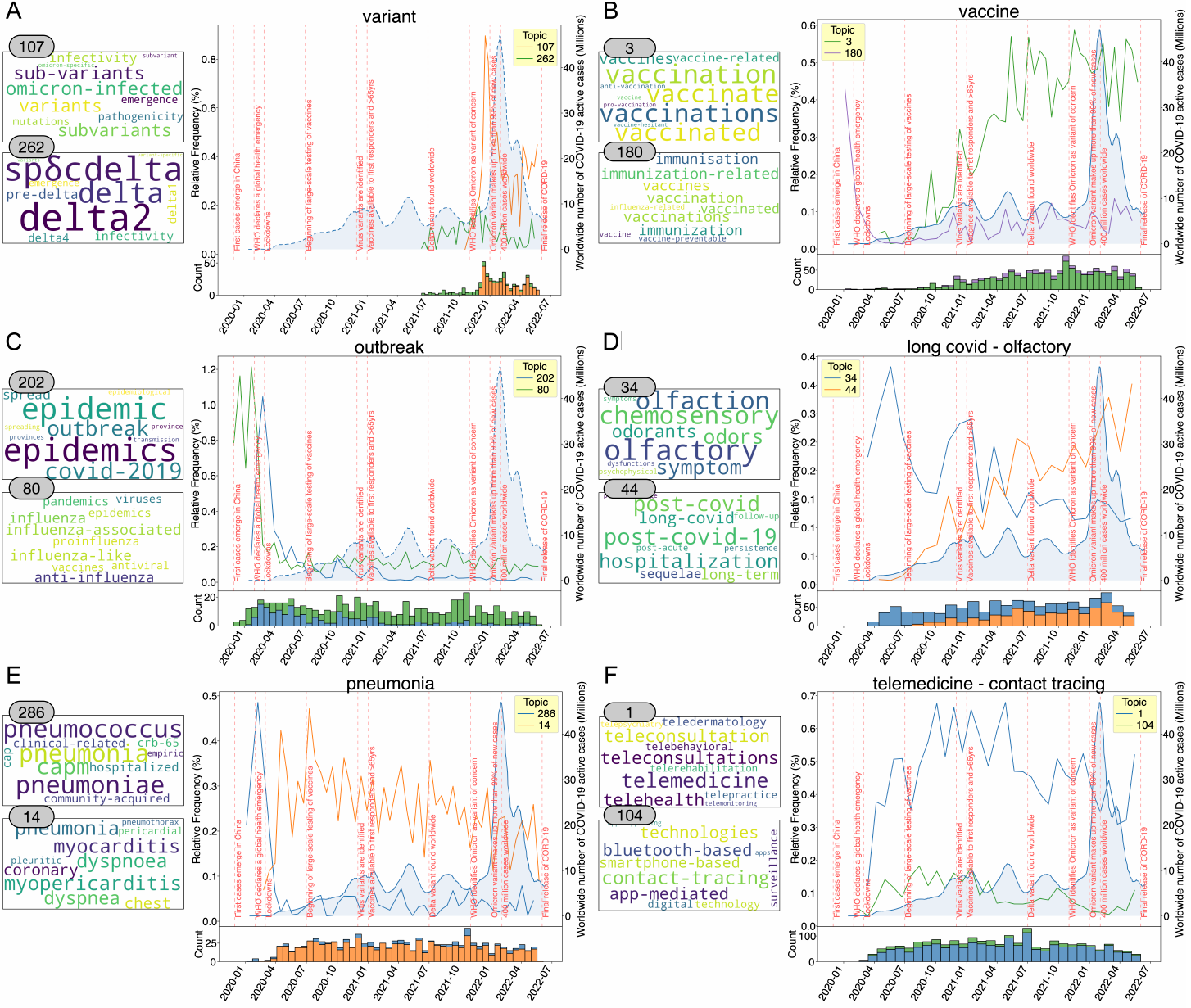} 
    \caption{{Visualizations of relevant example cases.} Each panel corresponds to the search of a keyword on CORToViz (see the title of plots).
    For each one, we show the word clouds generated for two topics and the line plots of the topics' time series.     
    (A) Variant: 
    topic on (sub)variants, among which omicron, whose spike anticipates a peak in active COVID cases shown in the background;
    topic on delta that increases when the variant spreads worldwide; 
    (B) Vaccine: 
    generic topic showing an increase in interest over time;
    immunization topic, more specific, with a similar trend.
    (C) Outbreak: 
    epidemic-related topic interesting at the beginning, but not much interesting after the first months;
    influenza, a topic with an early peak representing the large fraction of articles written on influenza prior to COVID, then less relevant and almost unrelated to COVID cases.
    (D) Olfactory and long covid: 
    the former peaking at the beginning of the pandemic and then decreasing;
    the latter showing a steadily increasing trend.
    (E) Pneumonia: 
    the first topic is decreasing in mid-2020 while the second topic, highlighting other co-morbidities, grows in interest.
    (F) Telemedicine and contact tracing:
    the first is steadily interesting;
    instead contact tracing is most interesting in the first months, but then loses interest (as it revealed hard to deploy in reality).
    }    
    \label{fig:examples}
\end{figure}

A broad spectrum of search sessions can be performed quickly and flexibly, as the system's average response time is about three seconds regardless of the number of bins. Fig.~\ref{fig:examples} reports a collection of use cases regarding topics that characterized the COVID-19 pandemic and discusses insights provided by the tool; plots use the default 2-week resolution.

\noindent
Panel (A) shows two interesting topics extracted by searching the word `variant'.  
\begin{itemize}
\item Topic-107 (focused on infectivity and pathogenicity of specific (sub)variants, among which omicron) appeared in December 2021 and had a dramatic rise peaking at the beginning of 2022, when related abstracts covered more than 8/1000 abstracts in CORD-19, with 55 in absolute numbers. 
Interestingly, we observe that topic-107's profile anticipates a peak of COVID-19 cases in the background, corresponding to the fourth wave of the pandemic. 
\item Topic-262 (focused on the delta variant, its sub-kinds, and infectivity effect) started between July and November 2021, with the rise of SARS-CoV-2 variants, and remained present throughout the pandemic.
\end{itemize}

\noindent
Panel (B) shows two topics extracted by searching the word `vaccine'.
\begin{itemize}
    \item Topic-3 (focused on vaccines and vaccinations) appears during the first months of the first lockdown period and starts increasing intensity at the beginning of large-scale vaccination testing; it further increases intensity until August-September 2021, then remains stable at 5/1000 abstracts for every bin (2 weeks). 
    \item Topic-180 (on immunization and immunization-related effects) is already present at the beginning of 2020, but at a very low intensity until January 2021, when vaccines become available in many countries~ \cite{whohistoryvaccines}; after that date, the topic increases its intensity until the end of the observation in June 2022.
\end{itemize}

\noindent
Panel (C) shows topics related to the `outbreak' keyword with very similar decreasing trends.
\begin{itemize}
    \item Topic-202 (on epidemic outbreaks) peaks at the beginning of the pandemic, when the World Health Organization declares COVID-19 as a global health emergency; then, it rapidly decreases.
    \item Topic-80 (on the influenza virus) peaks at the beginning of the pandemic, with more than 1/100 of abstracts per bi-weekly bin, an indication that before the pandemic outbreak, a relevant number of articles on viral species were focused on influenza. After the first six months, the topic relatively decreases and stabilizes at a lower intensity, reflecting the typical interest in influenza, rather independent of the course of the pandemic. 
\end{itemize}

\noindent
Panel (D) shows topics related to `olfactory' and `long covid'.
\begin{itemize}
    \item Topic-34 (olfactory) peaks at the beginning of the pandemic, when loss of odor sensing was first associated with COVID; it declines when this symptom becomes more known.
    \item Topic-44 (long covid) is present after the first pandemic wave in May 2020 and grows, at an intermittent rate, until the end of observation, where it presents the maximum relative frequency of 4/1000 abstracts in CORD-19, i.e., 46 articles.
\end{itemize}

\noindent
Panel (E) shows topics related to `pneumonia'. 
\begin{itemize}
    \item Topic-286 (highlighting central terms in pneumonia seen as a pulmonary disease) peaks at the beginning of the pandemic, when severe covid cases were associated with interstitial pneumonia. It then decreases, while the relevance of other clinical factors rises.
    \item Topic-14 (highlighting other aspects such as myocarditis and dyspnoea) specularly grows in interest, while the clinical models of severe COVID-19 link it to other co-morbidities.
\end{itemize}

\noindent
Panel (F) shows topics related to `telemedicine' and `contact tracing'.
\begin{itemize}
    \item Topic-1 (telemedicine) is quite relevant at all times of the pandemic, as the use of remote, domotic controls for monitoring a person's health has been a central theme throughout.
    \item Topic-104 (contact tracing) is less relevant, and it is possible to spot a decreasing trend, as the practical limitations of contact tracing became more evident. 
\end{itemize}

\subsection*{Topic trend comparison}

Table~\ref{tab:stat_test} describes 55 manually curated topics, grouped into 10 classes, which present interesting facets of the evolution of the pandemic. 

\begin{table}[ht]
    \centering
    \caption{Heatmap of trends of a manually-curated sample of 55 topics. For each topic, the median intensity ($\times$ 1000) has been computed for two-month intervals throughout the pandemic, as observed in CORD-19 (March 2020-June 2022). These values are represented by color nuances from white (lowest) to bright red (highest). To ease reading, we grouped topics with close semantics. Interest trends are either: increasing (e.g., vaccinations, variants, teleworking, long COVID), decreasing (e.g., outbreak topics, symptoms-related, testing and epidemiological modeling), or seasonal (e.g., tourism).}
        \resizebox{0.99\textwidth}{!}{%
\begin{tabular}{clrrrrrrrrrrrrrr}
 &  & \multicolumn{5}{c}{\textbf{2020}} & \multicolumn{6}{c}{\textbf{2021}} & \multicolumn{3}{c}{\textbf{2022}} \\
\cmidrule(r){3-7}
\cmidrule(r){8-13}
\cmidrule{14-16}
  \textbf{Topic ID} & \textbf{Description} & \textbf{1} & \textbf{2} & \textbf{3} & \textbf{4} & \textbf{5} & \textbf{6} & \textbf{7} & \textbf{8} & \textbf{9} & \textbf{10} & \textbf{11} & \textbf{12} & \textbf{13} & \textbf{14} \\
\midrule
\multicolumn{16}{c}{\textbf{Pandemic outbreak}}\\
174 & Italy                                                                 & {\cellcolor[HTML]{E83429}} 4,35 & {\cellcolor[HTML]{FCAE92}} 2,07 & {\cellcolor[HTML]{FCC4AD}} 1,61 & {\cellcolor[HTML]{FEDECF}} 1,06 & {\cellcolor[HTML]{FEE4D8}} 0,88 & {\cellcolor[HTML]{FDD3C1}} 1,29 & {\cellcolor[HTML]{FFF4EF}} 0,25 & {\cellcolor[HTML]{FFEFE8}} 0,45 & {\cellcolor[HTML]{FFF5F0}} 0,22 & {\cellcolor[HTML]{FFEEE6}} 0,50 & {\cellcolor[HTML]{FFF5F0}} 0,21 & {\cellcolor[HTML]{FFF2EC}} 0,32 & {\cellcolor[HTML]{FFF2EB}} 0,34 & {\cellcolor[HTML]{FFF4EE}} 0,28 \\
144 & Facemasks and Personal Protective Equipment                           & {\cellcolor[HTML]{E83429}} 1,45 & {\cellcolor[HTML]{FC8767}} 1,03 & {\cellcolor[HTML]{FC9474}} 0,96 & {\cellcolor[HTML]{FB7A5A}} 1,11 & {\cellcolor[HTML]{FB6D4D}} 1,17 & {\cellcolor[HTML]{ED392B}} 1,41 & {\cellcolor[HTML]{FFECE3}} 0,42 & {\cellcolor[HTML]{FCC4AD}} 0,71 & {\cellcolor[HTML]{FEE1D3}} 0,54 & {\cellcolor[HTML]{FFF2EB}} 0,36 & {\cellcolor[HTML]{FED8C7}} 0,59 & {\cellcolor[HTML]{FDD2BF}} 0,63 & {\cellcolor[HTML]{FEE8DE}} 0,45 & {\cellcolor[HTML]{FFF5F0}} 0,33 \\
320 & Ventilators                                                           & {\cellcolor[HTML]{FB6B4B}} 0,73 & {\cellcolor[HTML]{E83429}} 0,89 & {\cellcolor[HTML]{FEE7DB}} 0,32 & {\cellcolor[HTML]{FDCBB6}} 0,43 & {\cellcolor[HTML]{FCA183}} 0,56 & {\cellcolor[HTML]{FC8D6D}} 0,63 & {\cellcolor[HTML]{FCC4AD}} 0,45 & {\cellcolor[HTML]{FFF3ED}} 0,24 & {\cellcolor[HTML]{FFEFE8}} 0,27 & {\cellcolor[HTML]{FFEDE5}} 0,28 & {\cellcolor[HTML]{FEE2D5}} 0,35 & {\cellcolor[HTML]{FFF5F0}} 0,23 & {\cellcolor[HTML]{FEE5D9}} 0,33 & {\cellcolor[HTML]{FFF1EA}} 0,26 \\
279 & Management of waste and disinfection                                  & {\cellcolor[HTML]{E83429}} 1,27 & {\cellcolor[HTML]{FDD7C6}} 0,53 & {\cellcolor[HTML]{FC9C7D}} 0,82 & {\cellcolor[HTML]{FCBDA4}} 0,66 & {\cellcolor[HTML]{FFF5F0}} 0,30 & {\cellcolor[HTML]{FFF3ED}} 0,31 & {\cellcolor[HTML]{FDCCB8}} 0,58 & {\cellcolor[HTML]{FEDFD0}} 0,49 & {\cellcolor[HTML]{FDCEBB}} 0,57 & {\cellcolor[HTML]{FEDACA}} 0,52 & {\cellcolor[HTML]{FDD7C6}} 0,53 & {\cellcolor[HTML]{FCC1A8}} 0,65 & {\cellcolor[HTML]{FFF0E8}} 0,34 & {\cellcolor[HTML]{FFEEE6}} 0,37 \\
136 & Self testing                                                          & {\cellcolor[HTML]{FB6B4B}} 1,12 & {\cellcolor[HTML]{E83429}} 1,34 & {\cellcolor[HTML]{F24633}} 1,26 & {\cellcolor[HTML]{FDCEBB}} 0,66 & {\cellcolor[HTML]{FEE1D4}} 0,57 & {\cellcolor[HTML]{FCC4AD}} 0,72 & {\cellcolor[HTML]{FC997A}} 0,91 & {\cellcolor[HTML]{FCB398}} 0,79 & {\cellcolor[HTML]{FFF5F0}} 0,40 & {\cellcolor[HTML]{FEE3D6}} 0,56 & {\cellcolor[HTML]{FEDCCD}} 0,60 & {\cellcolor[HTML]{FDD3C1}} 0,64 & {\cellcolor[HTML]{FDC6B0}} 0,70 & {\cellcolor[HTML]{FEE3D6}} 0,56 \\

131 & Postponement and cancellation of surgeries and operations             & {\cellcolor[HTML]{FCB499}} 0,78 & {\cellcolor[HTML]{E83429}} 1,48 & {\cellcolor[HTML]{F34A36}} 1,36 & {\cellcolor[HTML]{FDC7B2}} 0,66 & {\cellcolor[HTML]{FB7555}} 1,14 & {\cellcolor[HTML]{FB7A5A}} 1,12 & {\cellcolor[HTML]{FC8F6F}} 0,99 & {\cellcolor[HTML]{FCB095}} 0,80 & {\cellcolor[HTML]{FEE3D6}} 0,48 & {\cellcolor[HTML]{FFF5F0}} 0,28 & {\cellcolor[HTML]{FDC6B0}} 0,67 & {\cellcolor[HTML]{FEE1D4}} 0,50 & {\cellcolor[HTML]{FCB095}} 0,80 & {\cellcolor[HTML]{FFF1EA}} 0,33 \\

11 & Surgeries, laparoscopies and endoscopies                               & {\cellcolor[HTML]{E83429}} 10,25 & {\cellcolor[HTML]{FEE3D6}} 3,04 & {\cellcolor[HTML]{FDD5C4}} 3,73 & {\cellcolor[HTML]{FEDFD0}} 3,32 & {\cellcolor[HTML]{FEE6DA}} 2,77 & {\cellcolor[HTML]{FFF2EC}} 1,83 & {\cellcolor[HTML]{FEEAE0}} 2,48 & {\cellcolor[HTML]{FFF0E8}} 2,02 & {\cellcolor[HTML]{FFEBE2}} 2,39 & {\cellcolor[HTML]{FFF4EE}} 1,70 & {\cellcolor[HTML]{FFEBE2}} 2,40 & {\cellcolor[HTML]{FEEAE0}} 2,47 & {\cellcolor[HTML]{FFECE4}} 2,29 & {\cellcolor[HTML]{FFF5F0}} 1,57 \\

\midrule
\multicolumn{16}{c}{\textbf{Understanding of the causes of severe disease}}\\
37 & Pneumonia and chest scans                                              & {\cellcolor[HTML]{E83429}} 7,25 & {\cellcolor[HTML]{FCAE92}} 3,55 & {\cellcolor[HTML]{FCC2AA}} 2,92 & {\cellcolor[HTML]{FDCBB6}} 2,58 & {\cellcolor[HTML]{FDCAB5}} 2,62 & {\cellcolor[HTML]{FED8C7}} 2,15 & {\cellcolor[HTML]{FEE1D3}} 1,82 & {\cellcolor[HTML]{FEE9DF}} 1,29 & {\cellcolor[HTML]{FFEEE7}} 1,00 & {\cellcolor[HTML]{FFF5F0}} 0,56 & {\cellcolor[HTML]{FFF4EF}} 0,60 & {\cellcolor[HTML]{FFEEE6}} 1,01 & {\cellcolor[HTML]{FFF0E8}} 0,90 & {\cellcolor[HTML]{FFF5F0}} 0,56 \\

173 & Hyperinflammation, cytokine storm, interleukin and tocilizumab        & {\cellcolor[HTML]{FFFFFF}} 0 & {\cellcolor[HTML]{FB6B4B}} 0,95 & {\cellcolor[HTML]{E83429}} 1,15 & {\cellcolor[HTML]{FB7757}} 0,89 & {\cellcolor[HTML]{FB7858}} 0,88 & {\cellcolor[HTML]{F34C37}} 1,06 & {\cellcolor[HTML]{FDD1BE}} 0,52 & {\cellcolor[HTML]{FCAB8F}} 0,68 & {\cellcolor[HTML]{FDD3C1}} 0,50 & {\cellcolor[HTML]{FFF5F0}} 0,28 & {\cellcolor[HTML]{FCAD90}} 0,67 & {\cellcolor[HTML]{FFF0E9}} 0,32 & {\cellcolor[HTML]{FEDFD0}} 0,45 & {\cellcolor[HTML]{FEE1D3}} 0,44 \\

196 & Cardiomyopathy, myocarditis, and cytokines                            & {\cellcolor[HTML]{FDC5AE}} 0,68 & {\cellcolor[HTML]{E83429}} 1,55 & {\cellcolor[HTML]{FC997A}} 0,96 & {\cellcolor[HTML]{FCC2AA}} 0,71 & {\cellcolor[HTML]{FCA689}} 0,88 & {\cellcolor[HTML]{FCA285}} 0,90 & {\cellcolor[HTML]{FFF2EC}} 0,29 & {\cellcolor[HTML]{FFF5F0}} 0,25 & {\cellcolor[HTML]{FEE9DF}} 0,40 & {\cellcolor[HTML]{FEE3D7}} 0,47 & {\cellcolor[HTML]{FEEAE0}} 0,39 & {\cellcolor[HTML]{FEDACA}} 0,55 & {\cellcolor[HTML]{FEE7DC}} 0,42 & {\cellcolor[HTML]{FFF2EC}} 0,29 \\

238 & Coagulopathies, thrombosis and thromboembolism                        & {\cellcolor[HTML]{FFFFFF}} 0 & {\cellcolor[HTML]{E83429}} 0,76 & {\cellcolor[HTML]{FA6648}} 0,65 & {\cellcolor[HTML]{F03D2D}} 0,73 & {\cellcolor[HTML]{FB7D5D}} 0,59 & {\cellcolor[HTML]{FCAD90}} 0,48 & {\cellcolor[HTML]{FFECE4}} 0,28 & {\cellcolor[HTML]{FCA689}} 0,49 & {\cellcolor[HTML]{FDC6B0}} 0,41 & {\cellcolor[HTML]{FFF5F0}} 0,24 & {\cellcolor[HTML]{F75B40}} 0,67 & {\cellcolor[HTML]{FED9C9}} 0,36 & {\cellcolor[HTML]{FCB296}} 0,47 & {\cellcolor[HTML]{FFEDE5}} 0,28 \\

34 & Symptomps: olfactory and chemosensory disfunctions                     & {\cellcolor[HTML]{FDC5AE}} 1,81 & {\cellcolor[HTML]{F5533B}} 3,08 & {\cellcolor[HTML]{E83429}} 3,41 & {\cellcolor[HTML]{FDD7C6}} 1,59 & {\cellcolor[HTML]{FCA285}} 2,22 & {\cellcolor[HTML]{FC8767}} 2,51 & {\cellcolor[HTML]{FDCBB6}} 1,74 & {\cellcolor[HTML]{FCC4AD}} 1,82 & {\cellcolor[HTML]{FEE1D4}} 1,44 & {\cellcolor[HTML]{FFF0E9}} 1,11 & {\cellcolor[HTML]{FDC7B2}} 1,77 & {\cellcolor[HTML]{FFF2EC}} 1,06 & {\cellcolor[HTML]{FFEFE8}} 1,14 & {\cellcolor[HTML]{FFF5F0}} 1,00 \\

35 & Manifestations of neuroinflammation, neuropathies and encephalitis     & {\cellcolor[HTML]{FFF5F0}} 0,78 & {\cellcolor[HTML]{FC8565}} 1,90 & {\cellcolor[HTML]{F4503A}} 2,30 & {\cellcolor[HTML]{F75C41}} 2,21 & {\cellcolor[HTML]{FC9576}} 1,76 & {\cellcolor[HTML]{E83429}} 2,52 & {\cellcolor[HTML]{EE3A2C}} 2,46 & {\cellcolor[HTML]{FCB89E}} 1,47 & {\cellcolor[HTML]{FC8A6A}} 1,86 & {\cellcolor[HTML]{FDD7C6}} 1,20 & {\cellcolor[HTML]{FCB398}} 1,51 & {\cellcolor[HTML]{FDD0BC}} 1,27 & {\cellcolor[HTML]{FCB89E}} 1,47 & {\cellcolor[HTML]{FCB095}} 1,54 \\

\midrule

\multicolumn{16}{c}{\textbf{Coronavirus severity and general traits}}\\

200 & COVID-19 infection in pregnancies: studies on preeclampsia            & {\cellcolor[HTML]{E83429}} 2,35 & {\cellcolor[HTML]{FDC6B0}} 0,96 & {\cellcolor[HTML]{FDC5AE}} 0,97 & {\cellcolor[HTML]{FEE9DF}} 0,51 & {\cellcolor[HTML]{FEE5D8}} 0,59 & {\cellcolor[HTML]{FEE9DF}} 0,50 & {\cellcolor[HTML]{FFF5F0}} 0,28 & {\cellcolor[HTML]{FEEAE0}} 0,49 & {\cellcolor[HTML]{FEE6DA}} 0,57 & {\cellcolor[HTML]{FFEEE7}} 0,40 & {\cellcolor[HTML]{FEEAE0}} 0,49 & {\cellcolor[HTML]{FEE8DD}} 0,53 & {\cellcolor[HTML]{FEE2D5}} 0,65 & {\cellcolor[HTML]{FFECE4}} 0,44 \\

49 & Relations with alcoholism and alcohol-associated conditions            & {\cellcolor[HTML]{FCBDA4}} 1,27 & {\cellcolor[HTML]{FFF5F0}} 0,53 & {\cellcolor[HTML]{FDD2BF}} 1,06 & {\cellcolor[HTML]{FCAE92}} 1,42 & {\cellcolor[HTML]{FDC6B0}} 1,18 & {\cellcolor[HTML]{FCBEA5}} 1,26 & {\cellcolor[HTML]{FCB296}} 1,38 & {\cellcolor[HTML]{FC9D7F}} 1,58 & {\cellcolor[HTML]{FC9879}} 1,62 & {\cellcolor[HTML]{FC9D7F}} 1,57 & {\cellcolor[HTML]{FC9D7F}} 1,57 & {\cellcolor[HTML]{F34C37}} 2,30 & {\cellcolor[HTML]{E83429}} 2,52 & {\cellcolor[HTML]{FC9777}} 1,63 \\

58 & Comorbidity of obesity and COVID-19                                    & {\cellcolor[HTML]{FCAE92}} 1,27 & {\cellcolor[HTML]{E83429}} 2,00 & {\cellcolor[HTML]{F24734}} 1,89 & {\cellcolor[HTML]{F44D38}} 1,85 & {\cellcolor[HTML]{F75C41}} 1,77 & {\cellcolor[HTML]{FCA98C}} 1,30 & {\cellcolor[HTML]{F75C41}} 1,77 & {\cellcolor[HTML]{ED392B}} 1,97 & {\cellcolor[HTML]{FC9272}} 1,44 & {\cellcolor[HTML]{FB7D5D}} 1,57 & {\cellcolor[HTML]{FEDFD0}} 0,94 & {\cellcolor[HTML]{FDCCB8}} 1,07 & {\cellcolor[HTML]{FFF5F0}} 0,68 & {\cellcolor[HTML]{FEE4D8}} 0,89 \\

44 & Long COVID and long-term post-COVID effects                            & {\cellcolor[HTML]{FFFFFF}} 0 & {\cellcolor[HTML]{FFF5F0}} 0,30 & {\cellcolor[HTML]{FFF5F0}} 0,30 & {\cellcolor[HTML]{FEE7DC}} 0,66 & {\cellcolor[HTML]{FEDECF}} 0,88 & {\cellcolor[HTML]{FEDACA}} 0,94 & {\cellcolor[HTML]{FDC6B0}} 1,24 & {\cellcolor[HTML]{FC997A}} 1,85 & {\cellcolor[HTML]{FC8F6F}} 1,99 & {\cellcolor[HTML]{FC9B7C}} 1,84 & {\cellcolor[HTML]{FC8262}} 2,16 & {\cellcolor[HTML]{FC8F6F}} 1,99 & {\cellcolor[HTML]{E83429}} 3,16 & {\cellcolor[HTML]{EC382B}} 3,09 \\

\midrule
\multicolumn{16}{c}{\textbf{Coronavirus and co-morbidities}}\\

71 & Comorbidity and mortality of cancer and malignancies                   & {\cellcolor[HTML]{FB7151}} 2,31 & {\cellcolor[HTML]{E83429}} 3,01 & {\cellcolor[HTML]{FCAB8F}} 1,57 & {\cellcolor[HTML]{FCAA8D}} 1,59 & {\cellcolor[HTML]{FCA183}} 1,71 & {\cellcolor[HTML]{FDD7C6}} 1,01 & {\cellcolor[HTML]{FCBDA4}} 1,35 & {\cellcolor[HTML]{FEE6DA}} 0,74 & {\cellcolor[HTML]{FDC5AE}} 1,24 & {\cellcolor[HTML]{FEE2D5}} 0,84 & {\cellcolor[HTML]{FEDFD0}} 0,90 & {\cellcolor[HTML]{FFF5F0}} 0,36 & {\cellcolor[HTML]{FFF1EA}} 0,47 & {\cellcolor[HTML]{FEE5D8}} 0,76 \\

230 & Interactions of COVID-19 with HER2 breast cancer                      & {\cellcolor[HTML]{FFFFFF}} 0 & {\cellcolor[HTML]{FED8C7}} 0,46 & {\cellcolor[HTML]{FFF5F0}} 0,32 & {\cellcolor[HTML]{FEE1D4}} 0,43 & {\cellcolor[HTML]{FEE0D2}} 0,44 & {\cellcolor[HTML]{FCBFA7}} 0,54 & {\cellcolor[HTML]{FDCAB5}} 0,51 & {\cellcolor[HTML]{FDCEBB}} 0,49 & {\cellcolor[HTML]{FEDCCD}} 0,45 & {\cellcolor[HTML]{FCB89E}} 0,56 & {\cellcolor[HTML]{FCA98C}} 0,60 & {\cellcolor[HTML]{FCAF93}} 0,58 & {\cellcolor[HTML]{E83429}} 0,91 & {\cellcolor[HTML]{F34935}} 0,86 \\

4 & Interactions of COVID-19 and diabetes and hyperglycemia                 & {\cellcolor[HTML]{FDD3C1}} 1,82 & {\cellcolor[HTML]{FFF5F0}} 0,79 & {\cellcolor[HTML]{FEDFD0}} 1,59 & {\cellcolor[HTML]{FCBCA2}} 2,32 & {\cellcolor[HTML]{FCB99F}} 2,35 & {\cellcolor[HTML]{FCB89E}} 2,39 & {\cellcolor[HTML]{FC9879}} 2,98 & {\cellcolor[HTML]{FCBEA5}} 2,25 & {\cellcolor[HTML]{FC8262}} 3,38 & {\cellcolor[HTML]{FC8161}} 3,41 & {\cellcolor[HTML]{FC8060}} 3,43 & {\cellcolor[HTML]{F4503A}} 4,27 & {\cellcolor[HTML]{EA362A}} 4,74 & {\cellcolor[HTML]{E83429}} 4,78 \\

220 & Vaccine-related and associated myocarditis                            & {\cellcolor[HTML]{FFFFFF}} 0 & {\cellcolor[HTML]{FFFFFF}} 0 & {\cellcolor[HTML]{FFFFFF}} 0 & {\cellcolor[HTML]{FFFFFF}} 0 & {\cellcolor[HTML]{FFFFFF}} 0 & {\cellcolor[HTML]{FFF5F0}} 0,26 & {\cellcolor[HTML]{FFF3ED}} 0,29 & {\cellcolor[HTML]{FFEEE6}} 0,38 & {\cellcolor[HTML]{FEE7DB}} 0,49 & {\cellcolor[HTML]{FDD3C1}} 0,71 & {\cellcolor[HTML]{FCBDA4}} 0,90 & {\cellcolor[HTML]{FDCDB9}} 0,76 & {\cellcolor[HTML]{FCAA8D}} 1,05 & {\cellcolor[HTML]{E83429}} 1,96 \\

248 & Correlation of COVID-19 and Kawasaki disease                          & {\cellcolor[HTML]{FFFFFF}} 0 & {\cellcolor[HTML]{FC9E80}} 0,52 & {\cellcolor[HTML]{FC9373}} 0,54 & {\cellcolor[HTML]{E83429}} 0,76 & {\cellcolor[HTML]{FC8F6F}} 0,55 & {\cellcolor[HTML]{FC8666}} 0,57 & {\cellcolor[HTML]{F75B40}} 0,68 & {\cellcolor[HTML]{F03F2E}} 0,73 & {\cellcolor[HTML]{FDC9B3}} 0,41 & {\cellcolor[HTML]{FCB095}} 0,47 & {\cellcolor[HTML]{FCAD90}} 0,48 & {\cellcolor[HTML]{FEE7DC}} 0,32 & {\cellcolor[HTML]{FFF5F0}} 0,25 & {\cellcolor[HTML]{FB7757}} 0,61 \\

52 & Interactions of Coccidioidomycosis, fungal infections and COVID-19     & {\cellcolor[HTML]{FFFFFF}} 0 & {\cellcolor[HTML]{FFF5F0}} 0,29 & {\cellcolor[HTML]{FEE6DA}} 0,65 & {\cellcolor[HTML]{FFF0E9}} 0,40 & {\cellcolor[HTML]{FEE1D3}} 0,78 & {\cellcolor[HTML]{FFECE4}} 0,50 & {\cellcolor[HTML]{FDC9B3}} 1,12 & {\cellcolor[HTML]{FCC4AD}} 1,18 & {\cellcolor[HTML]{FB7555}} 2,17 & {\cellcolor[HTML]{F6583E}} 2,51 & {\cellcolor[HTML]{E83429}} 2,92 & {\cellcolor[HTML]{F6553C}} 2,54 & {\cellcolor[HTML]{F03D2D}} 2,80 & {\cellcolor[HTML]{FB7151}} 2,22 \\

\midrule
\multicolumn{16}{c}{\textbf{Innovative treatments}}\\

23 & Hydroxychloroquine and cardiotoxic side effects                          & {\cellcolor[HTML]{FC8767}} 4,23 & {\cellcolor[HTML]{E83429}} 6,42 & {\cellcolor[HTML]{FC8D6D}} 4,09 & {\cellcolor[HTML]{FCA98C}} 3,32 & {\cellcolor[HTML]{FC9D7F}} 3,66 & {\cellcolor[HTML]{FDC5AE}} 2,52 & {\cellcolor[HTML]{FEE2D5}} 1,59 & {\cellcolor[HTML]{FEE8DE}} 1,25 & {\cellcolor[HTML]{FEE0D2}} 1,69 & {\cellcolor[HTML]{FFF1EA}} 0,80 & {\cellcolor[HTML]{FFF1EA}} 0,81 & {\cellcolor[HTML]{FFEEE7}} 0,93 & {\cellcolor[HTML]{FFF4EF}} 0,63 & {\cellcolor[HTML]{FFF5F0}} 0,56 \\

179 & Clinical studies on antihypertensives targeting angiotensin           & {\cellcolor[HTML]{FCBBA1}} 0,78 & {\cellcolor[HTML]{E83429}} 1,64 & {\cellcolor[HTML]{FCA082}} 0,96 & {\cellcolor[HTML]{FCA98C}} 0,90 & {\cellcolor[HTML]{FCAD90}} 0,88 & {\cellcolor[HTML]{FC9373}} 1,04 & {\cellcolor[HTML]{FEDBCC}} 0,55 & {\cellcolor[HTML]{FEE1D3}} 0,51 & {\cellcolor[HTML]{FEE5D8}} 0,46 & {\cellcolor[HTML]{FFF5F0}} 0,24 & {\cellcolor[HTML]{FFF5F0}} 0,25 & {\cellcolor[HTML]{FFF5F0}} 0,25 & {\cellcolor[HTML]{FEE5D9}} 0,45 & {\cellcolor[HTML]{FFEFE8}} 0,33 \\

13 & Treatments based on heterocyclic compounds                             & {\cellcolor[HTML]{FDD2BF}} 2,13 & {\cellcolor[HTML]{FFF5F0}} 1,40 & {\cellcolor[HTML]{FEE0D2}} 1,93 & {\cellcolor[HTML]{E83429}} 4,15 & {\cellcolor[HTML]{FDD5C4}} 2,09 & {\cellcolor[HTML]{FCBCA2}} 2,44 & {\cellcolor[HTML]{FEE5D8}} 1,82 & {\cellcolor[HTML]{FDCCB8}} 2,21 & {\cellcolor[HTML]{FC9B7C}} 2,89 & {\cellcolor[HTML]{FCBEA5}} 2,41 & {\cellcolor[HTML]{FCB398}} 2,56 & {\cellcolor[HTML]{FCA98C}} 2,70 & {\cellcolor[HTML]{FCC3AB}} 2,35 & {\cellcolor[HTML]{FEDECF}} 1,98 \\

22 & Therapies based on flavonoids and other phytochemicals                 & {\cellcolor[HTML]{F6553C}} 2,90 & {\cellcolor[HTML]{FFF5F0}} 0,59 & {\cellcolor[HTML]{FEE1D4}} 1,09 & {\cellcolor[HTML]{FC8D6D}} 2,21 & {\cellcolor[HTML]{FCBFA7}} 1,57 & {\cellcolor[HTML]{FCBFA7}} 1,57 & {\cellcolor[HTML]{FCB99F}} 1,65 & {\cellcolor[HTML]{FCC1A8}} 1,56 & {\cellcolor[HTML]{FC8969}} 2,27 & {\cellcolor[HTML]{F85F43}} 2,78 & {\cellcolor[HTML]{FC8161}} 2,36 & {\cellcolor[HTML]{E83429}} 3,28 & {\cellcolor[HTML]{F03D2D}} 3,16 & {\cellcolor[HTML]{F75B40}} 2,83 \\

8 & Efficacy of antimycobacterial and anti-TBC treatments                   & {\cellcolor[HTML]{FCA78B}} 2,78 & {\cellcolor[HTML]{FFF2EC}} 1,24 & {\cellcolor[HTML]{FFF5F0}} 1,15 & {\cellcolor[HTML]{FCB99F}} 2,49 & {\cellcolor[HTML]{FEE3D6}} 1,73 & {\cellcolor[HTML]{FDC6B0}} 2,26 & {\cellcolor[HTML]{FCAD90}} 2,68 & {\cellcolor[HTML]{FDCEBB}} 2,12 & {\cellcolor[HTML]{FCA98C}} 2,75 & {\cellcolor[HTML]{FC9576}} 3,06 & {\cellcolor[HTML]{FC8262}} 3,36 & {\cellcolor[HTML]{FB7B5B}} 3,49 & {\cellcolor[HTML]{EA362A}} 4,51 & {\cellcolor[HTML]{E83429}} 4,55 \\

334 & Corticosteroids in in-hospital treatment                              & {\cellcolor[HTML]{FFFFFF}} 0 & {\cellcolor[HTML]{FDD3C1}} 0,30 & {\cellcolor[HTML]{E83429}} 0,48 & {\cellcolor[HTML]{F96346}} 0,43 & {\cellcolor[HTML]{F75B40}} 0,44 & {\cellcolor[HTML]{F44D38}} 0,45 & {\cellcolor[HTML]{FC9070}} 0,38 & {\cellcolor[HTML]{FDCBB6}} 0,31 & {\cellcolor[HTML]{FB7353}} 0,41 & {\cellcolor[HTML]{FEE1D3}} 0,28 & {\cellcolor[HTML]{FC9373}} 0,37 & {\cellcolor[HTML]{FFEDE5}} 0,25 & {\cellcolor[HTML]{FFF5F0}} 0,23 & {\cellcolor[HTML]{FDD7C6}} 0,29 \\

24 & Opioids, medical and non-medical uses in the pandemic                  & {\cellcolor[HTML]{FC8262}} 2,21 & {\cellcolor[HTML]{FFF5F0}} 0,59 & {\cellcolor[HTML]{FDD1BE}} 1,28 & {\cellcolor[HTML]{FDC5AE}} 1,42 & {\cellcolor[HTML]{FCB296}} 1,66 & {\cellcolor[HTML]{FC9B7C}} 1,93 & {\cellcolor[HTML]{FB7D5D}} 2,27 & {\cellcolor[HTML]{FB6E4E}} 2,46 & {\cellcolor[HTML]{FC9576}} 1,99 & {\cellcolor[HTML]{FC9474}} 2,01 & {\cellcolor[HTML]{FC8767}} 2,16 & {\cellcolor[HTML]{FC7F5F}} 2,26 & {\cellcolor[HTML]{FC7F5F}} 2,26 & {\cellcolor[HTML]{E83429}} 3,08 \\

\midrule
\multicolumn{16}{c}{\textbf{SARS-CoV-2}}\\

150 & SARS-CoV-2 receptor bindings and ACE2                                 & {\cellcolor[HTML]{E83429}} 2,16 & {\cellcolor[HTML]{FFF5F0}} 0,36 & {\cellcolor[HTML]{FCC4AD}} 0,96 & {\cellcolor[HTML]{FEE0D2}} 0,71 & {\cellcolor[HTML]{FCA588}} 1,24 & {\cellcolor[HTML]{FCC4AD}} 0,97 & {\cellcolor[HTML]{FEE5D8}} 0,63 & {\cellcolor[HTML]{FCAD90}} 1,17 & {\cellcolor[HTML]{FDD5C4}} 0,81 & {\cellcolor[HTML]{FFEEE6}} 0,49 & {\cellcolor[HTML]{FDD5C4}} 0,81 & {\cellcolor[HTML]{FFF0E8}} 0,45 & {\cellcolor[HTML]{FFECE4}} 0,50 & {\cellcolor[HTML]{FEE9DF}} 0,56 \\
15 & Serology and immunoassays                                              & {\cellcolor[HTML]{FFF5F0}} 1,14 & {\cellcolor[HTML]{FC9E80}} 2,34 & {\cellcolor[HTML]{E83429}} 3,46 & {\cellcolor[HTML]{FB7151}} 2,84 & {\cellcolor[HTML]{F75C41}} 3,05 & {\cellcolor[HTML]{FC8464}} 2,63 & {\cellcolor[HTML]{F6553C}} 3,13 & {\cellcolor[HTML]{F24734}} 3,25 & {\cellcolor[HTML]{FC9879}} 2,42 & {\cellcolor[HTML]{FED9C9}} 1,67 & {\cellcolor[HTML]{FDD1BE}} 1,77 & {\cellcolor[HTML]{FCB89E}} 2,06 & {\cellcolor[HTML]{FED8C7}} 1,68 & {\cellcolor[HTML]{FEEAE1}} 1,36 \\
89 & Epitopes of antigens for SARS-CoV-2                                    & {\cellcolor[HTML]{FFFFFF}} 0 & {\cellcolor[HTML]{FFF5F0}} 0,64 & {\cellcolor[HTML]{FFF4EE}} 0,66 & {\cellcolor[HTML]{FCBDA4}} 0,95 & {\cellcolor[HTML]{FCA78B}} 1,03 & {\cellcolor[HTML]{FB6D4D}} 1,26 & {\cellcolor[HTML]{F5533B}} 1,34 & {\cellcolor[HTML]{E83429}} 1,45 & {\cellcolor[HTML]{FEDFD0}} 0,81 & {\cellcolor[HTML]{FB7A5A}} 1,20 & {\cellcolor[HTML]{FC8F6F}} 1,12 & {\cellcolor[HTML]{F6583E}} 1,33 & {\cellcolor[HTML]{FDD7C6}} 0,84 & {\cellcolor[HTML]{FCA285}} 1,05 \\
299 & Phenotyope, genome and polymorphisms of SARS-CoV-2                    & {\cellcolor[HTML]{FFFFFF}} 0 & {\cellcolor[HTML]{FC8262}} 0,56 & {\cellcolor[HTML]{FEE9DF}} 0,32 & {\cellcolor[HTML]{FEE6DA}} 0,33 & {\cellcolor[HTML]{FB7D5D}} 0,57 & {\cellcolor[HTML]{F7593F}} 0,64 & {\cellcolor[HTML]{FC8767}} 0,55 & {\cellcolor[HTML]{FC8B6B}} 0,54 & {\cellcolor[HTML]{FFF3ED}} 0,28 & {\cellcolor[HTML]{E83429}} 0,72 & {\cellcolor[HTML]{FFF5F0}} 0,27 & {\cellcolor[HTML]{FEE8DE}} 0,32 & {\cellcolor[HTML]{FCAE92}} 0,47 & {\cellcolor[HTML]{FC8262}} 0,56 \\

124 & Variants and substitutions                                            & {\cellcolor[HTML]{FFFFFF}} 0 & {\cellcolor[HTML]{FFF2EC}} 0,31 & {\cellcolor[HTML]{FFF1EA}} 0,32 & {\cellcolor[HTML]{FFF5F0}} 0,28 & {\cellcolor[HTML]{FFF4EE}} 0,29 & {\cellcolor[HTML]{FC7F5F}} 1,01 & {\cellcolor[HTML]{E83429}} 1,36 & {\cellcolor[HTML]{FA6648}} 1,13 & {\cellcolor[HTML]{EB372A}} 1,34 & {\cellcolor[HTML]{F6563D}} 1,20 & {\cellcolor[HTML]{FA6849}} 1,12 & {\cellcolor[HTML]{FCA285}} 0,82 & {\cellcolor[HTML]{FEE3D7}} 0,46 & {\cellcolor[HTML]{FCC2AA}} 0,66 \\

262 & Delta variant                                                         & {\cellcolor[HTML]{FFFFFF}} 0 & {\cellcolor[HTML]{FFFFFF}} 0 & {\cellcolor[HTML]{FFFFFF}} 0 & {\cellcolor[HTML]{FFFFFF}} 0 & {\cellcolor[HTML]{FFFFFF}} 0 & {\cellcolor[HTML]{FFFFFF}} 0 & {\cellcolor[HTML]{FFFFFF}} 0 & {\cellcolor[HTML]{FFFFFF}} 0 & {\cellcolor[HTML]{FFF5F0}} 0,37 & {\cellcolor[HTML]{FCA689}} 0,83 & {\cellcolor[HTML]{E83429}} 1,33 & {\cellcolor[HTML]{FCA082}} 0,87 & {\cellcolor[HTML]{FDC6B0}} 0,69 & {\cellcolor[HTML]{FFEEE6}} 0,44 \\
107 & Omicron variant, subvariants and infectivity                          & {\cellcolor[HTML]{FFFFFF}} 0 & {\cellcolor[HTML]{FFFFFF}} 0 & {\cellcolor[HTML]{FFFFFF}} 0 & {\cellcolor[HTML]{FFFFFF}} 0 & {\cellcolor[HTML]{FFFFFF}} 0 & {\cellcolor[HTML]{FFFFFF}} 0 & {\cellcolor[HTML]{FFFFFF}} 0 & {\cellcolor[HTML]{FFFFFF}} 0 & {\cellcolor[HTML]{FFFFFF}} 0 & {\cellcolor[HTML]{FFFFFF}} 0 & {\cellcolor[HTML]{FFF5F0}} 0,20 & {\cellcolor[HTML]{E83429}} 4,51 & {\cellcolor[HTML]{FB7C5C}} 3,15 & {\cellcolor[HTML]{F6563D}} 3,86 \\
\midrule

\multicolumn{16}{c}{\textbf{Vaccination}}\\

59 & Infection-prevention campaigns and adherence                           & {\cellcolor[HTML]{FFF5F0}} 0,73 & {\cellcolor[HTML]{FDC6B0}} 1,34 & {\cellcolor[HTML]{FEDACA}} 1,15 & {\cellcolor[HTML]{FCA78B}} 1,62 & {\cellcolor[HTML]{F44F39}} 2,36 & {\cellcolor[HTML]{E83429}} 2,58 & {\cellcolor[HTML]{FC997A}} 1,74 & {\cellcolor[HTML]{FDD0BC}} 1,25 & {\cellcolor[HTML]{FDC5AE}} 1,35 & {\cellcolor[HTML]{FEE7DC}} 0,97 & {\cellcolor[HTML]{FDCBB6}} 1,30 & {\cellcolor[HTML]{FEDCCD}} 1,13 & {\cellcolor[HTML]{FEEAE1}} 0,91 & {\cellcolor[HTML]{FFEFE8}} 0,84 \\
81 & Available vaccines, immunization and immunogenicity                    & {\cellcolor[HTML]{E83429}} 1,86 & {\cellcolor[HTML]{FFF5F0}} 0,52 & {\cellcolor[HTML]{FEE2D5}} 0,76 & {\cellcolor[HTML]{FDD4C2}} 0,86 & {\cellcolor[HTML]{FFEFE8}} 0,59 & {\cellcolor[HTML]{F5533B}} 1,67 & {\cellcolor[HTML]{FCC2AA}} 0,99 & {\cellcolor[HTML]{FCB095}} 1,10 & {\cellcolor[HTML]{FCA082}} 1,21 & {\cellcolor[HTML]{FA6648}} 1,57 & {\cellcolor[HTML]{FC8A6A}} 1,35 & {\cellcolor[HTML]{FC8E6E}} 1,32 & {\cellcolor[HTML]{FCAD90}} 1,13 & {\cellcolor[HTML]{FCC3AB}} 0,98 \\
215 & mRNA-based drugs and vaccines                                         & {\cellcolor[HTML]{FB7252}} 0,73 & {\cellcolor[HTML]{FFFFFF}} 0 & {\cellcolor[HTML]{FEDBCC}} 0,42 & {\cellcolor[HTML]{FDCAB5}} 0,48 & {\cellcolor[HTML]{FFF5F0}} 0,29 & {\cellcolor[HTML]{FC8767}} 0,67 & {\cellcolor[HTML]{FCA689}} 0,58 & {\cellcolor[HTML]{FCC3AB}} 0,50 & {\cellcolor[HTML]{FCB296}} 0,55 & {\cellcolor[HTML]{FCBCA2}} 0,52 & {\cellcolor[HTML]{E83429}} 0,90 & {\cellcolor[HTML]{FA6547}} 0,77 & {\cellcolor[HTML]{F6583E}} 0,80 & {\cellcolor[HTML]{FB7D5D}} 0,70 \\
3 & Vaccination: for, against and hesitant                                  & {\cellcolor[HTML]{FFFFFF}} 0 & {\cellcolor[HTML]{FFF4EE}} 0,38 & {\cellcolor[HTML]{FFF5F0}} 0,32 & {\cellcolor[HTML]{FEEAE0}} 0,81 & {\cellcolor[HTML]{FEE2D5}} 1,16 & {\cellcolor[HTML]{FCBEA5}} 2,09 & {\cellcolor[HTML]{FC9576}} 2,98 & {\cellcolor[HTML]{FB7757}} 3,68 & {\cellcolor[HTML]{F96346}} 4,10 & {\cellcolor[HTML]{F14331}} 4,72 & {\cellcolor[HTML]{E83429}} 5,07 & {\cellcolor[HTML]{ED392B}} 4,93 & {\cellcolor[HTML]{F44F39}} 4,51 & {\cellcolor[HTML]{EF3C2C}} 4,87 \\

\midrule
\multicolumn{16}{c}{\textbf{Social aspects}}\\

27 & Coping with loneliness and disconnection for elderly people            & {\cellcolor[HTML]{FFFFFF}} 0 & {\cellcolor[HTML]{FFF5F0}} 0,95 & {\cellcolor[HTML]{FEE7DC}} 1,09 & {\cellcolor[HTML]{FFF0E9}} 1,00 & {\cellcolor[HTML]{FEE2D5}} 1,14 & {\cellcolor[HTML]{FDCDB9}} 1,26 & {\cellcolor[HTML]{FCBBA1}} 1,36 & {\cellcolor[HTML]{E83429}} 2,03 & {\cellcolor[HTML]{FC9B7C}} 1,53 & {\cellcolor[HTML]{F14331}} 1,95 & {\cellcolor[HTML]{FCA082}} 1,50 & {\cellcolor[HTML]{FCA082}} 1,50 & {\cellcolor[HTML]{FCA588}} 1,47 & {\cellcolor[HTML]{FC9D7F}} 1,52 \\
39 & Stress and burnout of healthcare workers                               & {\cellcolor[HTML]{FDCAB5}} 1,49 & {\cellcolor[HTML]{FFF4EE}} 1,10 & {\cellcolor[HTML]{FEE5D9}} 1,28 & {\cellcolor[HTML]{FFF5F0}} 1,07 & {\cellcolor[HTML]{FCA689}} 1,73 & {\cellcolor[HTML]{FDD2BF}} 1,43 & {\cellcolor[HTML]{FB7656}} 2,03 & {\cellcolor[HTML]{FC8060}} 1,97 & {\cellcolor[HTML]{FCAB8F}} 1,69 & {\cellcolor[HTML]{FC8A6A}} 1,90 & {\cellcolor[HTML]{FA6547}} 2,13 & {\cellcolor[HTML]{FC8161}} 1,96 & {\cellcolor[HTML]{FCB99F}} 1,60 & {\cellcolor[HTML]{E83429}} 2,41 \\
137 & Caregivers                                                            & {\cellcolor[HTML]{FB7A5A}} 1,13 & {\cellcolor[HTML]{FCA98C}} 0,84 & {\cellcolor[HTML]{FDC6B0}} 0,65 & {\cellcolor[HTML]{FC8262}} 1,07 & {\cellcolor[HTML]{FCA285}} 0,87 & {\cellcolor[HTML]{E83429}} 1,51 & {\cellcolor[HTML]{FC9474}} 0,96 & {\cellcolor[HTML]{FCBBA1}} 0,73 & {\cellcolor[HTML]{FCB79C}} 0,76 & {\cellcolor[HTML]{FDD4C2}} 0,56 & {\cellcolor[HTML]{FEE8DD}} 0,39 & {\cellcolor[HTML]{FCA98C}} 0,84 & {\cellcolor[HTML]{FFF5F0}} 0,23 & {\cellcolor[HTML]{FDD2BF}} 0,57 \\

149 & Work from home                                                        & {\cellcolor[HTML]{FC9070}} 0,78 & {\cellcolor[HTML]{FFF3ED}} 0,30 & {\cellcolor[HTML]{FFF0E9}} 0,31 & {\cellcolor[HTML]{FFF5F0}} 0,28 & {\cellcolor[HTML]{FDCBB6}} 0,54 & {\cellcolor[HTML]{FFECE4}} 0,35 & {\cellcolor[HTML]{FCAA8D}} 0,68 & {\cellcolor[HTML]{FC8E6E}} 0,80 & {\cellcolor[HTML]{FC8E6E}} 0,80 & {\cellcolor[HTML]{FCA285}} 0,71 & {\cellcolor[HTML]{FC8B6B}} 0,81 & {\cellcolor[HTML]{EE3A2C}} 1,11 & {\cellcolor[HTML]{EB372A}} 1,13 & {\cellcolor[HTML]{E83429}} 1,14 \\

267 & Effects of restrictions on travelling, traffic, commuting             & {\cellcolor[HTML]{FFFFFF}} 0 & {\cellcolor[HTML]{FFFFFF}} 0 & {\cellcolor[HTML]{FDCCB8}} 0,39 & {\cellcolor[HTML]{FFF4EE}} 0,27 & {\cellcolor[HTML]{FFEFE8}} 0,29 & {\cellcolor[HTML]{EC382B}} 0,66 & {\cellcolor[HTML]{FDD4C2}} 0,37 & {\cellcolor[HTML]{FFF5F0}} 0,26 & {\cellcolor[HTML]{F03D2D}} 0,65 & {\cellcolor[HTML]{EF3C2C}} 0,66 & {\cellcolor[HTML]{FCAE92}} 0,45 & {\cellcolor[HTML]{E83429}} 0,67 & {\cellcolor[HTML]{FCA285}} 0,47 & {\cellcolor[HTML]{FCA183}} 0,47 \\

316 & Travelling and tourism                                                & {\cellcolor[HTML]{FFFFFF}} 0 & {\cellcolor[HTML]{EF3C2C}} 0,56 & {\cellcolor[HTML]{FEDACA}} 0,32 & {\cellcolor[HTML]{FC9777}} 0,43 & {\cellcolor[HTML]{E83429}} 0,57 & {\cellcolor[HTML]{FDCAB5}} 0,35 & {\cellcolor[HTML]{FC8666}} 0,45 & {\cellcolor[HTML]{FCA486}} 0,41 & {\cellcolor[HTML]{FFF3ED}} 0,25 & {\cellcolor[HTML]{F03F2E}} 0,56 & {\cellcolor[HTML]{FFEFE8}} 0,27 & {\cellcolor[HTML]{FFF5F0}} 0,25 & {\cellcolor[HTML]{EF3C2C}} 0,56 & {\cellcolor[HTML]{FC9070}} 0,44 \\

5 & e-Learning: teaching, education and university                          & {\cellcolor[HTML]{FFF5F0}} 0,68 & {\cellcolor[HTML]{FEE8DE}} 1,07 & {\cellcolor[HTML]{FDD2BF}} 1,57 & {\cellcolor[HTML]{FCBDA4}} 1,93 & {\cellcolor[HTML]{FCA588}} 2,30 & {\cellcolor[HTML]{FCA285}} 2,35 & {\cellcolor[HTML]{F6563D}} 3,50 & {\cellcolor[HTML]{F85D42}} 3,41 & {\cellcolor[HTML]{F96044}} 3,36 & {\cellcolor[HTML]{FB7555}} 3,06 & {\cellcolor[HTML]{F75B40}} 3,43 & {\cellcolor[HTML]{FC9474}} 2,56 & {\cellcolor[HTML]{E83429}} 4,00 & {\cellcolor[HTML]{F96044}} 3,37 \\

\midrule
\multicolumn{16}{c}{\textbf{Technology}}\\

1 & Telemedicine, teleconsultations, and telehealth                         & {\cellcolor[HTML]{FFF5F0}} 1,86 & {\cellcolor[HTML]{FC8161}} 4,65 & {\cellcolor[HTML]{FB7555}} 4,90 & {\cellcolor[HTML]{FB6E4E}} 5,03 & {\cellcolor[HTML]{F6583E}} 5,42 & {\cellcolor[HTML]{EB372A}} 6,00 & {\cellcolor[HTML]{FB7151}} 4,96 & {\cellcolor[HTML]{E83429}} 6,08 & {\cellcolor[HTML]{FCA285}} 3,98 & {\cellcolor[HTML]{FC9879}} 4,18 & {\cellcolor[HTML]{FA6849}} 5,16 & {\cellcolor[HTML]{FC9E80}} 4,06 & {\cellcolor[HTML]{FCAA8D}} 3,83 & {\cellcolor[HTML]{FCAB8F}} 3,79 \\

104 & Smartphone-based contact tracing technologies                         & {\cellcolor[HTML]{E83429}} 1,73 & {\cellcolor[HTML]{FCA689}} 1,09 & {\cellcolor[HTML]{FB7252}} 1,40 & {\cellcolor[HTML]{FC7F5F}} 1,32 & {\cellcolor[HTML]{FC9879}} 1,18 & {\cellcolor[HTML]{FB6C4C}} 1,43 & {\cellcolor[HTML]{F96245}} 1,49 & {\cellcolor[HTML]{FFF5F0}} 0,50 & {\cellcolor[HTML]{FCC4AD}} 0,92 & {\cellcolor[HTML]{FDCBB6}} 0,88 & {\cellcolor[HTML]{FDC7B2}} 0,90 & {\cellcolor[HTML]{FFF2EB}} 0,54 & {\cellcolor[HTML]{FEE5D9}} 0,68 & {\cellcolor[HTML]{FCC3AB}} 0,93 \\

\midrule
\multicolumn{16}{c}{\textbf{Pandemic models}}\\

20 & Estimations of mortality and fatality                                  & {\cellcolor[HTML]{FDD4C2}} 2,17 & {\cellcolor[HTML]{E83429}} 4,20 & {\cellcolor[HTML]{FDCCB8}} 2,30 & {\cellcolor[HTML]{FCB296}} 2,65 & {\cellcolor[HTML]{FCA78B}} 2,77 & {\cellcolor[HTML]{FC9D7F}} 2,90 & {\cellcolor[HTML]{FCBBA1}} 2,53 & {\cellcolor[HTML]{FEDECF}} 2,04 & {\cellcolor[HTML]{FFF4EE}} 1,53 & {\cellcolor[HTML]{FDD1BE}} 2,23 & {\cellcolor[HTML]{FFF4EF}} 1,50 & {\cellcolor[HTML]{FFF5F0}} 1,48 & {\cellcolor[HTML]{FFF4EF}} 1,51 & {\cellcolor[HTML]{FEE8DE}} 1,80 \\
224 & Prediction and prognosis of COVID-19                                  & {\cellcolor[HTML]{FB7050}} 0,78 & {\cellcolor[HTML]{FCB79C}} 0,56 & {\cellcolor[HTML]{FFF1EA}} 0,32 & {\cellcolor[HTML]{FC9D7F}} 0,64 & {\cellcolor[HTML]{F5533B}} 0,86 & {\cellcolor[HTML]{E83429}} 0,96 & {\cellcolor[HTML]{FB7151}} 0,78 & {\cellcolor[HTML]{FDD7C6}} 0,45 & {\cellcolor[HTML]{FFEBE2}} 0,35 & {\cellcolor[HTML]{FCB89E}} 0,56 & {\cellcolor[HTML]{FEDCCD}} 0,43 & {\cellcolor[HTML]{FEE5D8}} 0,39 & {\cellcolor[HTML]{FCB499}} 0,57 & {\cellcolor[HTML]{FFF5F0}} 0,29 \\
50 & Epidemiological modelling and simulations                              & {\cellcolor[HTML]{FB7C5C}} 2,54 & {\cellcolor[HTML]{E83429}} 3,42 & {\cellcolor[HTML]{F34935}} 3,16 & {\cellcolor[HTML]{FC9B7C}} 2,13 & {\cellcolor[HTML]{FDD0BC}} 1,43 & {\cellcolor[HTML]{FCAB8F}} 1,91 & {\cellcolor[HTML]{FCB99F}} 1,74 & {\cellcolor[HTML]{FEE5D9}} 1,06 & {\cellcolor[HTML]{FEE8DE}} 0,97 & {\cellcolor[HTML]{FEE7DC}} 1,00 & {\cellcolor[HTML]{FEE5D8}} 1,07 & {\cellcolor[HTML]{FFF5F0}} 0,64 & {\cellcolor[HTML]{FEE5D9}} 1,05 & {\cellcolor[HTML]{FEE3D7}} 1,10 \\
319 & Zoonotic and human-to-animal interactions                             & {\cellcolor[HTML]{FFFFFF}} 0 & {\cellcolor[HTML]{FDD4C2}} 0,33 & {\cellcolor[HTML]{FEDFD0}} 0,31 & {\cellcolor[HTML]{FFF1EA}} 0,26 & {\cellcolor[HTML]{FEE7DB}} 0,29 & {\cellcolor[HTML]{FEDFD0}} 0,31 & {\cellcolor[HTML]{FEE6DA}} 0,29 & {\cellcolor[HTML]{FFF5F0}} 0,24 & {\cellcolor[HTML]{F75C41}} 0,54 & {\cellcolor[HTML]{FC8464}} 0,47 & {\cellcolor[HTML]{E83429}} 0,60 & {\cellcolor[HTML]{FC8767}} 0,47 & {\cellcolor[HTML]{FC8F6F}} 0,45 & {\cellcolor[HTML]{F34C37}} 0,56 \\
72 & Air-quality, anthropogenic pollution and aerosol                       & {\cellcolor[HTML]{FC9D7F}} 1,36 & {\cellcolor[HTML]{FFF5F0}} 0,55 & {\cellcolor[HTML]{FFF0E8}} 0,63 & {\cellcolor[HTML]{FDCEBB}} 0,99 & {\cellcolor[HTML]{FCBCA2}} 1,14 & {\cellcolor[HTML]{FC8161}} 1,56 & {\cellcolor[HTML]{FEE7DB}} 0,76 & {\cellcolor[HTML]{FDC6B0}} 1,06 & {\cellcolor[HTML]{FDCCB8}} 1,01 & {\cellcolor[HTML]{FC997A}} 1,39 & {\cellcolor[HTML]{FCB499}} 1,20 & {\cellcolor[HTML]{FC9576}} 1,42 & {\cellcolor[HTML]{E83429}} 2,10 & {\cellcolor[HTML]{FC8969}} 1,52 \\
\bottomrule
\end{tabular}

    }
    \label{tab:stat_test}
\end{table}

\noindent
{\bf Pandemic outbreak.} 
The first group includes topics that were very popular (high frequency) during the pandemic outbreak, then lost interest. Among them, the outbreak in Italy -- the first European country hit by the virus \cite{capobianchi2020molecular,cerqua2022did} -- and the lack of preparedness in terms of organization or medical, protective, and testing equipment. Still, in the early period, the burden of COVID-19 caused several postponements and cancellations of surgeries and operations, with deleterious effects on other pathologies.

\noindent
{\bf Understanding the causes of severe disease}.
At the beginning of the pandemic, clinicians were struggling to understand the causes of the pandemic, originally considered a pulmonary disease, and later on better classified as a vascular-inflammatory disease. 
Red areas illustrate well the progression of the understanding process, throughout the first year of the pandemic (2020). 
We note that the neuroinflammation and neuropathies, related to the long-term effects of COVID-19, are slightly delayed, compared to the other topics in this group.

\noindent
{\bf Coronavirus severity and general traits.} 
Effects of general traits (pregnancies, alcoholism, obesity) were studied throughout the pandemic. 
At a given point, the term {\it long Covid} started to denote symptoms that persist after the end of the acute COVID-19 disease.

\noindent
{\bf Coronavirus and co-morbidities.}
Important studies relate COVID-19 severity to 
cancer, 
breast cancer, 
diabetes and hyperglycemia, 
and other more specific
comorbidities such as myocarditis, 
Kawasaki disease, and 
fungal infections.

\noindent
{\bf Innovative treatments.}
Throughout the pandemic, a number of innovative treatments were tried, with quite controversial effects. Among them, are chloroquine, angiotensin, heterocyclic compounds, flavonoids, and antimycobacterial treatments. 
Other classical treatments, such as corticosteroids and opioids, were evaluated throughout the pandemic. 

\noindent
{\bf SARS-CoV-2.} 
During the pandemic, virologists have studied the SARS-CoV-2 virus and its evolution; some terms relate to immunological aspects (reception bindings, serology, epitopes),
other to the evolution of the genome, including variants, and specifically some important variants such as Delta and Omicron; of course, these terms are created at the time when variants first appear.

\noindent
{\bf Vaccination.}
The interest in these topics grew after the end of 2020 when the first vaccines became available to the public for mass immunization. Discussions on vaccination constantly increase until the end of the study.

\noindent
{\bf Social aspects.}
Aspects related to the mental health of caregivers rise after the first year of the pandemic, then remain at high intensity. Similarly, changes in lifestyle imposed by the lockdowns, such as the work-from-home policies, show an increasing trend. 
Tourism and traveling exhibit a seasonal trend, with several equally spaced peaks. The effects of distance learning were also studied, including the impact on children and adolescents, such as lack of socialization and physical activity.

\noindent
{\bf Technology.}
Telemedicine and telehealth, made necessary by the criticality of health systems, show an increase in intensity, as many studies were devoted to the benefits induced by telemedicine in COVID-19.
Instead, technologies for contact tracing, which seemed promising at the beginning of the pandemic, had an initial peak and then a decreasing trend.

\noindent 
{\bf Pandemic models.}
Several communities studied the pandemic from a variety of viewpoints, including epidemiological models, virological models, zoonotic models that focus on interactions with animal species, and environmental models (principally on pollution); these exhibit different peaks of intensity all over the pandemic.

\section*{Applying the paradigm to a novel domain}
The full-stack process described in this research is virtually applicable to any corpus of medium-to-large-sized textual documents, using any topic model of choice, and a time-series visualizer. The main challenge stands in the data collection phase, in charge of gathering data on other domains of interest.

\begin{figure}[h!]
    \centering
    \includegraphics[width=\linewidth]{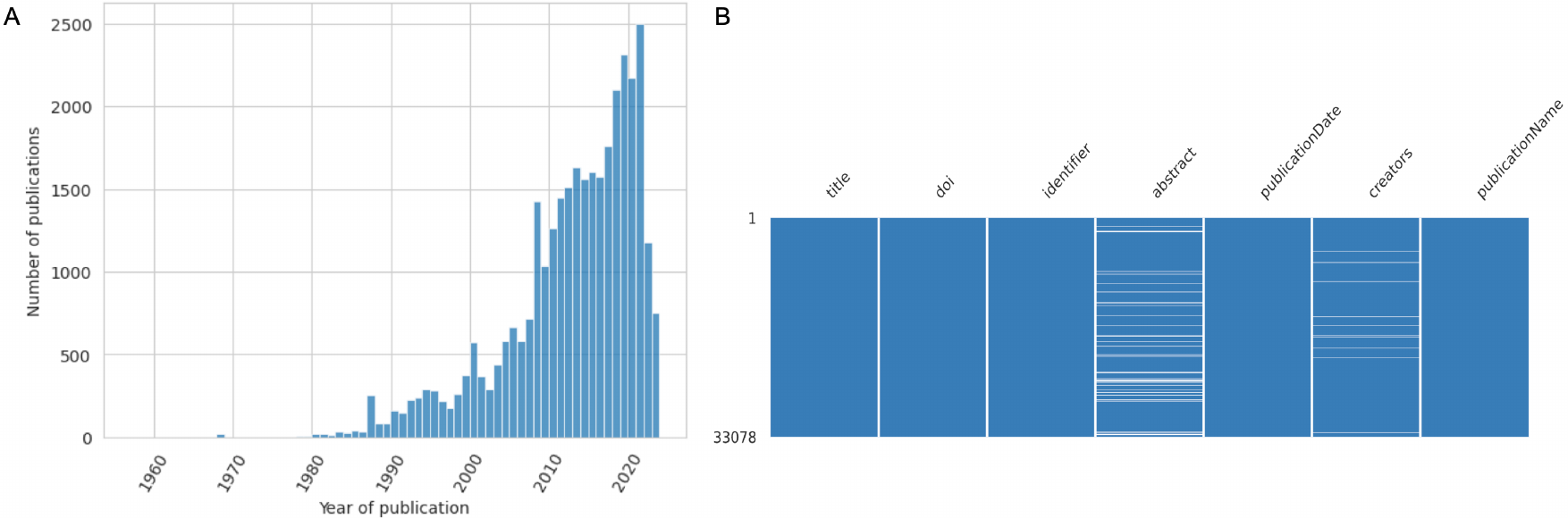}
    \caption{{Visualizations of the exploratory analysis of the dataset on climate change from Springer Nature Group.} 
    (A) Yearly number of publications. The number is below 500 articles per year until 2008. Then, it ramps to 1,500 articles per year for about one decade. In the last 5 years, the number exceeds 2,000 articles.
    (B) Data-density display of seven metadata fields for the whole dataset, which shows, in general, the good quality of the dataset. We note some unavailable data in the abstract and creators fields.
    } 
    \label{fig:climate}
\end{figure}

To demonstrate the applicability of the pipeline, we conducted a data collection activity targeting climate change-related scientific literature. 
To this end, we exploited the public endpoints exposed by the Springer Nature publishing group. We obtained a dataset of 33,723 scientific abstracts, upon which we performed an exploratory analysis of records and their metadata---see Fig.~\ref{fig:climate}, whose panel (A) shows the significant increase in the volume of scientific publications about climate change in the last 40 years. 
By using the preparation stage of the CORToViz pipeline, 
we obtained a final dataset of 29,886 
abstracts (after language selection).
Then, the self-tuning data pipeline developed for CORToViz (Fig.~\ref{fig:architecture}) was reapplied on the abstracts to obtain a new topic model, that includes 166 topics; these could be then explored with dashboards similar to CORToViz; 
note that our work required minimal adaptation and was performed in about two days by the first author.

\section{Discussion}
By critically analyzing the scientific abstracts of the CORD-19 dataset, 
this research has shed light on key factors of the evolution of research questions on COVID-19, SARS-CoV-2, and the whole pandemic phenomenon. 
The proposed technological pipeline and associated visualizer combine state-of-the-art methods, targeting both data extraction efficiency and user-friendly topic extraction and visualization with integrated statistical testing. Through the lens of scientific research, CORToViz enables the understanding of
individual aspects (topics) that characterized the COVID-19 crisis worldwide, with their interactions and timing. 
Additionally, we also showcased the benefit of having a statistical approach for dynamic topic modeling built on the results of deep learning-based language models.

For the purpose of this work, we have chosen the most comprehensive dataset (CORD-19) available to date, without the need for integrating different datasets. 
Progress of this research could consider extending the COVID-19 corpus with other corpora targeting the last year of the pandemic. In particular, LitCOVID is still actively curated at the time of writing. Unfortunately, it is
focused on PubMed, thus it covers a smaller fraction of research articles w.r.t. CORD-19; it excludes non-medical articles, which are instead an important portion of CORD-19. 


In perspective, an approach like the one of CORToViz can be applied to both highly technical texts (e.g., scientific research abstracts) and general texts (e.g., book reviews). The flexibility of the approach makes it applicable to very diverse domains and markets; it enables improved access to summarized and digested content both on domain-specific Web content (e.g., reviews on water-scoping machines) and more domain-general ones (e.g., climate change). Our contribution enables addressing the needs of stakeholders requiring a one-click stack, providing immediate high-level analytics to quickly grasp trends, for instance in e-commerce reviews, events feedback tweets, public engagement threads, or any other business in which observing temporal trends is crucial.

Since the methods embedded in the modeling pipeline and in the dashboard are not specific to COVID-19 and SARS-CoV-2, the CORD-19 topic extraction pipeline and visualizer are easily adapted to literature repositories with a similar organization. To demonstrate this aspect, we 
employed the public API of the Springer Nature Group to retrieve publications 
about climate change, and then we quickly applied the full pipeline to build a dataset, organized by topic, fully compatible with our topic visualizer.

\vspace{2mm}
\noindent
\textbf{Data and code availability statement}
The original CORD-19 dataset is available at the GitHub repository of the project, at the URL \url{https://github.com/allenai/cord19}. 
Both data processing pipeline and application are available as Docker images on \url{https://hub.docker.com/r/frinve/cortoviz/}.
The CORToViz application is freely available on \url{http://gmql.eu/cortoviz}.

\vspace{2mm}
\noindent
\textbf{Authorship contribution statement.}
F.I. and A.B. conceived the work;
F.I., A.B., and S.C. jointly conceptualized and designed the framework; 
S.C. insisted on making the framework reusable for other domains; 
F.I. selected a coherent set of up-to-date technologies and developed the core pipeline;
F.I. and A.B. curated the user experience; 
F.I. drafted the manuscript, A.B. and S.C. improved it; 
all authors revised the final version of the manuscript;
S.C. supervised the project.

\vspace{2mm}
\noindent
\textbf{Declaration of Competing Interest.}
The authors declare that they have no known competing financial interests or personal relationships that could have appeared to influence the work reported in this paper.


\vspace{2mm}
\noindent
\textbf{Funding.}
This research is supported 
by the ERC AdG 693174 project ``GeCo: data-driven Genomic Computing";
by the PNRR-PE-AI FAIR project funded by the NextGenerationEU program; and 
by the TEThYS project within NGI Search funded by the European Commission, as part of the Horizon Europe Research and Innovation Programme under Grant Agreement Nº101069364.


\printcredits

\bibliographystyle{cas-model2-names}

\bibliography{document}



\end{document}